# ClinReadNet: A clinical reading-inspired network for low-dose abdominal CT image quality assessment

Xianye Xiao[1,2], Yulong Zou[1,2], Yujie Luo[1,2], Taihui Yu[3], Cun-Jing Zheng[3], Yuan-ming Geng[4], Shuihua Wang[5], Yudong Zhang[6], Jin Hong[2,*]

1. School of Mathematics and Computer Sciences, Nanchang University, Nanchang 330031, China
2. School of Information Engineering, Nanchang University, Nanchang 330031, China
3. Department of Radiology, Sun Yat-Sen Memorial Hospital, Sun Yat-Sen University, Guangzhou 510120, China
4. Department of Stomatology, Zhujiang Hospital, Southern Medical University, Guangzhou 510282, China
5. Department of Biological Sciences, School of Science, Xi'an Jiaotong Liverpool University, Suzhou 215123, China
6. School of Computer Science and Engineering, Southeast University, Nanjing 210096, China

E-mail: 9109223042@email.ncu.edu.cn; 9109223090@email.ncu.edu.cn; 9109223056@email.ncu.edu.cn; yuth3@mail.sysu.edu.cn; zhengcj@mail2.sysu.edu.cn; gym@smu.edu.cn; Shuihua.Wang@xjtlu.edu.cn; yudongzhang@seu.edu.cn; hongjin@ncu.edu.cn;

* Correspondence should be addressed to Jin Hong

**Abstract:** In abdominal CT imaging, optimizing the balance between radiation dose and image quality is crucial, and the primary prerequisite is accurate image quality assessment. Clinical practice uses doctors' subjective judgment as the gold standard, but it is time-consuming and costly; therefore, developing a low-dose, no-reference image quality assessment (No-reference IQA) model that mimics doctors' reading habits for evaluating CT image quality has significant practical value. This paper proposes a novel deep learning-based framework, ClinReadNet, whose design aligns with the clinical reading logic of radiologists: first, it introduces the Sobel ordinal quality network (SOQN) module, which can simultaneously focus on edge details highly relevant to image quality and the quality distribution pattern of the entire image, accurately matching the clinical image-reading judgment habit of "considering both local details and overall context"; second, the framework integrates the (shifted) window multi-scale temperature multi-head self-attention ((S)W-MTMSA) module, which further replicates the radiologists' image-reading process of shifting from overall scanning to local focusing, and accurately locks in regions of interest through multi-sharpness attention; third, it designs the hierarchical ranked probability score (HRPS) loss function, which combines the dual logics of coarse classification and fine classification, while paying attention to the distance information between grading labels, effectively improving the performance of image quality assessment. Experiments conducted on the LDCTIQAG2023 dataset show that the proposed method achieves the current state-of-the-art (SOTA) performance: the values of Pearson’s linear correlation coefficient (PLCC), Spearman’s rank-order correlation coefficient (SROCC), and Kendall’s rank-order correlation coefficient (KROCC) reach 0.9507, 0.9554, and 0.8629 respectively, with the sum of their absolute values (Score) being 2.7690, outperforming existing methods.

## 1. Introduction

CT scans rely on X-rays for imaging, and X-rays are a form of ionizing radiation—a characteristic that gives rise to the core contradiction between "radiation dose and image quality" in CT applications [1]. On one hand, CT scans with relatively high radiation doses may increase the risk of long-term cancer development in examinees; this health hazard of cumulative radiation exposure is particularly prominent for patients who require multiple follow-up scans [2]. On the other hand, if X-ray output is excessively reduced to lower the radiation dose, it will lead to a significant increase in noise and artifacts in CT images [3]. This, in turn, may obscure subtle lesions or cause blurring of lesion boundaries, interfering with doctors' judgment on the nature of lesions and ultimately compromising the accuracy of clinical diagnosis.

Therefore, in the process of CT imaging, it is necessary to minimize radiation dose to the greatest extent while ensuring that CT images meet the requirements for accurate clinical diagnosis. This has led to increasing attention to image quality in the field of medical imaging, and the use of effective image quality assessment (IQA) methods to evaluate CT images has become increasingly essential.

For IQA, clinical practice still regards physicians' subjective judgment as the gold standard [4]. However, physician-based assessment is time-consuming and costly [5]. In this context, a series of IQA methods have been proposed. These methods are generally categorized into three types: full-reference (FR), reduced-reference (RR), and no-reference (NR) approaches [6]. Among them, two widely used Image Quality Assessment (IQA) performance metrics, Peak Signal to Noise Ratio (PSNR) and Structural Similarity Index (SSIM) [7] , are often used to quantify the performance of FR-IQ methods. Both FR-IQA and RR-IQA methods rely on reference images to conduct assessments. In real-world clinical scenarios, however, obtaining corresponding reference images is often challenging. Therefore, NR-IQA methods have become the most suitable option for clinical settings [5].

Based on differences in technical principles, mainstream NR-IQA methods can be divided into two major categories: traditional non-deep learning methods and deep learning-based methods. Typical examples of traditional non-deep learning methods include the natural image quality evaluator (NIQE) [8] and the blind/referenceless image spatial quality evaluator (BRISQUE) [9]. These methods are designed based on the statistical laws of natural images. However, medical images differ significantly from natural images in terms of anatomical structures and noise characteristics, making pre-defined statistical models unable to accurately match the quality degradation features of medical images, which often leads to deviations in assessment results[10, 11]. Deep learning-based methods have shown great potential in the field of medical image analysis[12-15] and have also made progress in abdominal CT image quality assessment[16]. Yet, they still have limitations: first, most schemes remain confined to borrowing improved ideas from image classification tasks to enhance quality grading performance; second, most existing studies fail to focus on edge features that are highly relevant to image quality and also fail to fully attend to the unique ordinal nature of medical image grading, without constructing targeted model architectures around this core characteristic; third, relevant research has not attempted to derive design inspiration from the clinical image reading logic of radiologists, resulting in a disconnect between the model's assessment logic and actual clinical interpretation habits, which can hinder the clinical adoption and trustworthiness of such models among physicians.

To address the aforementioned limitations, we propose a novel abdominal CT image quality assessment framework named ClinReadNet. Our method aligns with the routine image-reading logic of radiologists: by establishing a collaborative learning mechanism for detailed and global information, we simulate the radiologists' observation process of "first grasping the overall quality from a macro perspective and then examining local details from a micro perspective"; by enhancing the extraction of edge features, we conform to the radiologists' focus on the clarity of organ boundaries and the distinguishability of tissue layers; by optimizing model training using the progressive gradient characteristic of quality labels, we match the radiologists' subjective judgment logic of "quality grades transitioning gradually from excellent to poor".

Specifically, we adopt Swin Transformer as the basic backbone of the model, regard the IQA task as a classification task first, and then innovatively design the Sobel ordinal quality network (SOQN) module in the shallow layer of the network: this module enhances the accurate extraction of edge details through the Sobel operator, while fusing global quality grade information, realizing the collaborative extraction of features from two dimensions (local structure and macro quality) and providing a solid feature foundation for subsequent quality assessment. In addition, in view of the multi-scale characteristics of abdominal CT image quality features, we propose the (shifted) window multi-scale temperature multi-head self-attention ((S)W-MTMSA) module, which can capture coarse, medium-grained, and fine quality features respectively. Through an adaptive weight adjustment mechanism, it maintains consistency with the radiologists' "macro-to-micro" image-reading logic. Furthermore, to optimize the model training process, we design the hierarchical ranked probability score (HRPS) loss function: this loss function adopts a "two-layer evaluation" structure, first conducting an overall preliminary assessment of image quality, and then performing refined calibration of quality scores, whose logic is highly compatible with application scenarios in clinical practice. The contributions of this study are as follows:

(i) We propose a novel abdominal CT image quality assessment framework, ClinReadNet, which establishes a clinical-reading-inspired paradigm by synergistically integrating the SOQN, (S)W-MTMSA, and HRPS modules. This paradigm explicitly models the radiologist's workflow—from global context grasping to local detail examination, and from coarse screening to fine scoring—thereby significantly enhancing the performance of the model.

(ii) We developed a SOQN module, which explicitly encodes the "considering both local details and overall context" logic by jointly processing edge structures and ordinal quality distributions.

(iii) We constructed a (S)W-MTMSA module. The (S)W-MTMSA module operationalizes the "macro-to-micro" reading logic through a novel multi-temperature attention mechanism, enabling adaptive focus switching that mimics radiologists' observation patterns.

(iv) We designed a hierarchical sorting probability score (HRPS) loss function to embody the "coarse-to-fine" diagnostic judgment process in its hierarchical structure, ensuring the model's predictions respect the ordinal nature of quality grades in a clinically plausible manner.

(v) Our framework achieves excellent performance on the benchmark dataset: on the LDCTIQAG2023 dataset, it obtains the current state-of-the-art (SOTA) Pearson’s linear correlation coefficient (PLCC) of 0.9570, Spearman’s rank-order correlation coefficient (SROCC) of 0.9554, Kendall’s rank-order correlation coefficient (KROCC) of 0.8629, and overall correlation score (comprehensive value of all correlation coefficients) of 2.7690.

## 2. Related work

### 2.1. Medical image grading

In recent years, the field of medical image grading has developed rapidly and in depth. It is different from ordinary image classification tasks - grading tasks focus on "hierarchical judgment" and cover two core directions: disease grading and image quality assessment.

In disease grading research, diverse technical approaches have offered abundant solutions for automated grading: For diabetic retinopathy (DR) grading, Huang et al. [17] proposed a lesion-based self-supervised contrastive learning framework that encourages feature extractors to learn highly discriminative representations via contrastive prediction tasks with various data augmentations. Cheng [18] proposed the sparse range-constrained learning (SRCL) algorithm, innovatively integrating sparse representation search and image grading objectives into a single function, which improved accuracy in image grading. Khawaldeh et al. [19] applied an improved AlexNet to brain medical image classification (healthy, low-grade/high-grade tumors) for enhanced representation performance. Shazuli and Saravanan [20] proposed the IWOADL-RFIGR method, leveraging lightweight CNN for similarity measurement and least squares support vector machine (LS-SVM) for retinal fundus image classification, with hyperparameter optimization via the improved whale optimization algorithm (IWOA).

In image quality assessment research, scholars have proposed organ-specific optimization schemes: Han and Baek [21] designed a three-layer CNN-based anthropomorphic model observer for breast anatomical CT image assessment. Lauermann et al. [22] applied a deep learning algorithm (DLA) for automatic quality assessment of optical coherence tomography angiography (OCTA) images. Gao et al. [23] developed a closed-loop "automated annotation-deep learning" framework to realize end-to-end no-reference CT image quality assessment (applicable to abdominal and chest CT). Qi et al. [24] proposed a multi-module CNN-based fully automated framework for whole-body [18F]FDG PET/CT image quality assessment, generating standardized ROIs and outputting subjective scores and objective indicators such as SUVmax and SNR.

Overall, despite the multi-technical path exploration in current medical image grading research, most studies still rely on image classification improvement ideas to enhance performance and lack targeted modeling for the core "ordered labels" characteristic. Our work achieves collaborative modeling of global quality level information and local edge information, combined with a loss function featuring a two-layer evaluation structure, which aligns with the ordered label characteristic of medical image grading and fully exploits the value of the task's ordered nature.

**2.2. Abdominal CT image quality assessment**

The release of the LDCTIQAG2023 dataset has effectively promoted the emergence of no-reference assessment methods for abdominal CT image quality [16]. Unlike previous datasets that focused on CT images affected by a single type of artifact, this dataset not only covers various complex noises and artifacts in CT images but also includes perceptual scores provided by expert radiologists. Researchers have proposed no-reference methods from different perspectives to solve the task of abdominal CT image quality assessment.

For research on selecting suitable backbone for tasks: CHILL@UK The team adjusted the input-output layer of EfficientNetV2 [25] for mean square error loss regression and achieved an overall correlation score of 2.6719 on the LDCTIQAG2023 dataset [16]. The agaldran team used Swin Transformer [26] and BiT ResNeXt-50 [27] as independent backbones, combined with multi-head classifiers [28] and a joint loss of cross entropy loss and ranking probability loss [29, 30], achieving a total loss of 2.7427 on the dataset [16]. The RPI-AXIS team proposed a multi-dimensional attention network[31] with patch weighted dual branch prediction and mean square error loss , with a score of 2.6843[16]. In addition, Gong et al. [32] developed the DL-MO method through transfer learning, which integrates pre trained CNN, PLS-DA models, and internal noise components to optimize feature representation.

For multi-stage task adaptive models: Team Epoch proposed a two-stage multi-task (regression and classification) learning model, which extracts noisy images in the first stage and fuses them with the original image for quality evaluation in the second stage of regression and classification dual output. The model scored 2.6202 on LDCTIQAG2023 [16]. For clinical-oriented optimization, Lee et al. [9] proposed D2IQA, an innovative self-supervised object detection-based method that calculates CT image quality via virtual geometric object detection. It shows robustness to dose changes, with performance exceeding that of most no-reference metrics and competing with full-reference ones.

For multi-dimensional feature extraction models: The gabybaldleon team fused Vision Transformer[32] with convolutional networks and adopted a "teacher-student" training process, scoring 2.5671 on LDCTIQAG2023 [16]. FeatureNet designed a three-branch architecture, including Vision Transformer[32], Haralick texture extraction[33], wavelet convolution, to capture global pixel correlations, texture, and frequency-domain features [16], reaching 2.6550 on the dataset [16].

Despite progress in the quality assessment technology of abdominal CT images, existing models still have shortcomings: when fusing local and global features, the task characteristics are not fully integrated, such as local feature extraction ignoring the clinical significance of edges, global feature extraction not utilizing task hierarchy, and not drawing on clinical image reading logic, making it difficult to gain the trust of doctors and patients.

To address these limitations, the SOQN module proposed in this study differs significantly from existing methods in three aspects: a) The Sobel Branch focuses on edge features, directly reflecting core quality indicators of abdominal CT, rather than relying on indirect features like texture and frequency domain; b) The Ordinal Quality Branch integrates global features, suppresses noise, and quantifies quality into 21 ordered levels; c) The two branches adopt dynamic weighted fusion, improving the effectiveness of representation.

Meanwhile, the (S)W-MTMSA module proposed in this study also differs significantly from existing methods in three aspects: a) it combines global observation and local examination while flexibly adjusting the focus; b) it synchronously captures coarse, medium, and fine sharpness attention through multi-temperature attention; c) it performs adaptive weight adjustment on the multi-sharpness attention of different windows.

### 2.3. Evaluation metrics for ordinal tasks

In ordinal classification tasks, existing metrics are categorized by their focus on "hard classification results" or "probability prediction information". For hard classification outputs, key metrics reflecting error severity include quadratic-weighted kappa [34] and expected cost [35] . Quadratic-weighted kappa constructs a category distance-squared weight matrix to quantify prediction-true consistency and penalize severe errors. Expected cost links classification errors to actual losses via domain expert-defined cost matrices for clinical relevance. However, both metrics only focus on final category decisions and ignore probability prediction uncertainty.

With the development of model uncertainty quantification, probability prediction evaluation has emerged as a new direction. Proper Scoring Rules (PSRs) [36] are widely adopted for guiding true probability distribution output, including the Brier score [37] and the logarithmic score. The Brier score quantifies prediction accuracy via squared errors between predicted probability vectors and true one-hot labels, with good differentiability for training. The logarithmic score is negative logarithm, rewarding high confidence in correct categories as a local rule. However, neither accounts for inter-category ordinal relationships in ordinal classification, failing to distinguish between different error severities and thus unable to meet ordinal scenario evaluation needs.

To address the probability prediction evaluation gap in ordinal classification, the ranked probability score (RPS) [30] was proposed as a proper scoring rule. It integrates category order into probability evaluation by calculating the squared distance between predicted and true cumulative probability distributions, penalizing larger deviations more heavily.

In abdominal CT image quality assessment, some studies converted RPS into a loss function to leverage label ordinality, avoiding the "equal treatment of all errors" flaw of traditional loss functions and guiding reliable quality grade probability distributions. However, it lacks clinical scenario-oriented improvements.

This study further proposes the hierarchical ranked probability score (HRPS) loss function, which differs significantly from existing methods in three key aspects: a) It draws on radiologists' image-reading logic of "first overall screening, then detailed scoring"; b) The HRPS loss function is designed with a two-layer ranked probability score structure; c) It introduces dynamic normalization and entropy regularization terms to balance the loss scales of different samples and grades.

## 3. Method

### 3.1. Overall architecture of ClinReadNet

The overall framework of ClinReadNet is shown in **Fig. 1**, which consists of three main components: Sobel ordinal quality network (SOQN) module, (shifted) window multi-scale temperature multi-head self-attention module((S)W-MTMSA), and hierarchical ranked probability score (HRPS) loss. For the input abdominal CT images, the network adopts a structured processing flow to provide technical support for efficiently completing image quality assessment tasks.

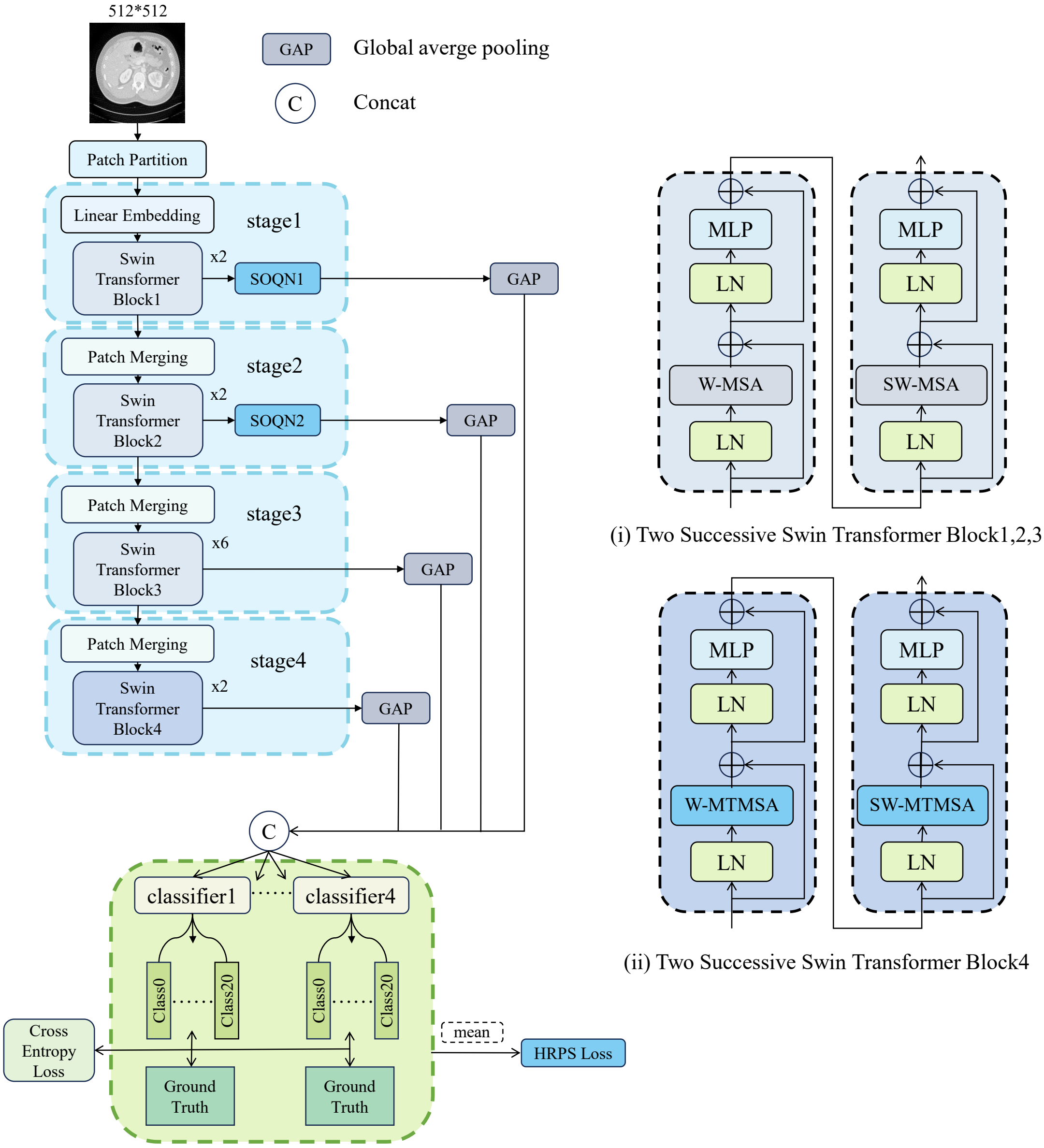

**Fig. 1** The overall framework of ClinReadNet.

Swin-T is adopted as the backbone network in this study, primarily motivated by its reliable performance in abdominal CT image quality assessment tasks. Specifically, the input image undergoes a preprocessing process before entering the feature extraction stage. In the feature extraction stage, the four stages of Swin-T process the passing feature maps in sequence. Among them, the feature maps output by stage 1 and stage 2 need to be additionally sent to the SOQN module for further processing. The SOQN module adopts a dual-branch design, including a Sobel branch and an ordinal quality branch: The Sobel branch performs convolution operations on the feature maps using Sobel kernels in four different directions, and completes feature fusion after enhancement processing; The ordinal quality branch integrates global features, suppresses noise interference, and quantifies image quality into 21 ordered levels by combining the hierarchical characteristics of labels, thereby enhancing the global quality modeling capability. Finally, the feature maps of the two branches are fused through a weighted method. Stage 3 retains the structure of the classic Swin transformer block unchanged, while in the Swin transformer block of Stage 4, W-MSA and SW-MSA are replaced with W-MTMSA and SW-MTMSA respectively. These two modules use a multi-temperature attention mechanism to capture multi-grained quality features, and rely on a quality-aware modulator to achieve adaptive adjustment of attention weights. The feature maps output by stage 3 and stage 4 do not require additional processing. Subsequently, the feature maps output by stage 1, stage 2, stage 3, and stage 4 are respectively subjected to global average pooling operations. The pooled feature vectors are concatenated and then input to four independent classification heads. The category weights of each classification head are different. Specifically, all category numbers are first divided into parts equal to the number of classification heads, and each part is assigned to one head. For each head, the weights of the categories belonging to that head are set as the number of classification heads, and the weights of the remaining categories are set as the reciprocal of the number of classification heads.[28] This approach aims to address issues such as gradient convergence and lack of diversity in predictions that easily occur when multi-header models adopt the same weight for each category in their loss calculation[28].

Finally, the final result of image quality assessment is obtained by averaging the predictions of the four classification heads. During the training process, the HRPS Loss and cross-entropy loss function are adopted as the joint loss function. The cross-entropy loss is calculated separately for each classification head, and the outputs of all classification heads are averaged across the classification head dimension. Then calculate the HRPS loss to directly optimize the performance of the average results of the four classification heads, ensuring that the predictive ability of the final multi-head average output is optimal. Take the average of the results of cross entropy loss and HRPS loss as the total loss to update and optimize the model parameters.

### 3.2. Sobel ordinal quality network (SOQN)

The architecture of the SOQN module is shown in **Fig. 2**, and its specific process is as follows (Note the feature maps should be reshaped from [H, W, C] to [C, H, W] before being input into the SOQN module): First, the input feature map $X$ undergoes normalization processing to improve numerical stability. Subsequently, the feature map sequentially passes through the channel attention and spatial attention modules to obtain the more discriminative preprocessed feature $X_{pre}$. Next, $X_{pre}$ is sent to two parallel branches: the Sobel branch and the ordinal quality branch. And then, the fusion weights for the two branches are then generated. The specific approach is: concatenate the outputs $S_{out}$ and $O_{out}$ of the two branches in the channel dimension. During training, gumbel softmax[38] is used on the concatenated features to generate weights, thereby improving the discriminability and differentiability of the branch weights. During validation and test, softmax is used to ensure output stability. Gumbel softmax[38] introduces Gumbel noise to enable differentiable discrete sampling for training, while softmax directly outputs deterministic probabilities for stable inference. The branch weights are used to perform weighted fusion on the features of the two branches to obtain the fused feature $F_{fused}$. This can be expressed by Equation (1).

$$w = \begin{cases} \sigma_1(f_{fuse}(F_{cat})), training \\ \sigma_2(f_{fuse}(F_{cat})), validation\ and\ test \end{cases} \tag{1}$$

where $F_{cat}$ represents the concatenated output of the two branches, $\sigma_1(\cdot)$ represents gumbel softmax, $\sigma_2(\cdot)$ represents softmax, $f_{fuse}(\cdot)$ represents the branch fusion module, including 1×1 convolution, batch normalization, activation function Mish, adaptive average pooling layer, 1×1 convolution, batch normalization, activation function Mish, and 1×1 convolution.

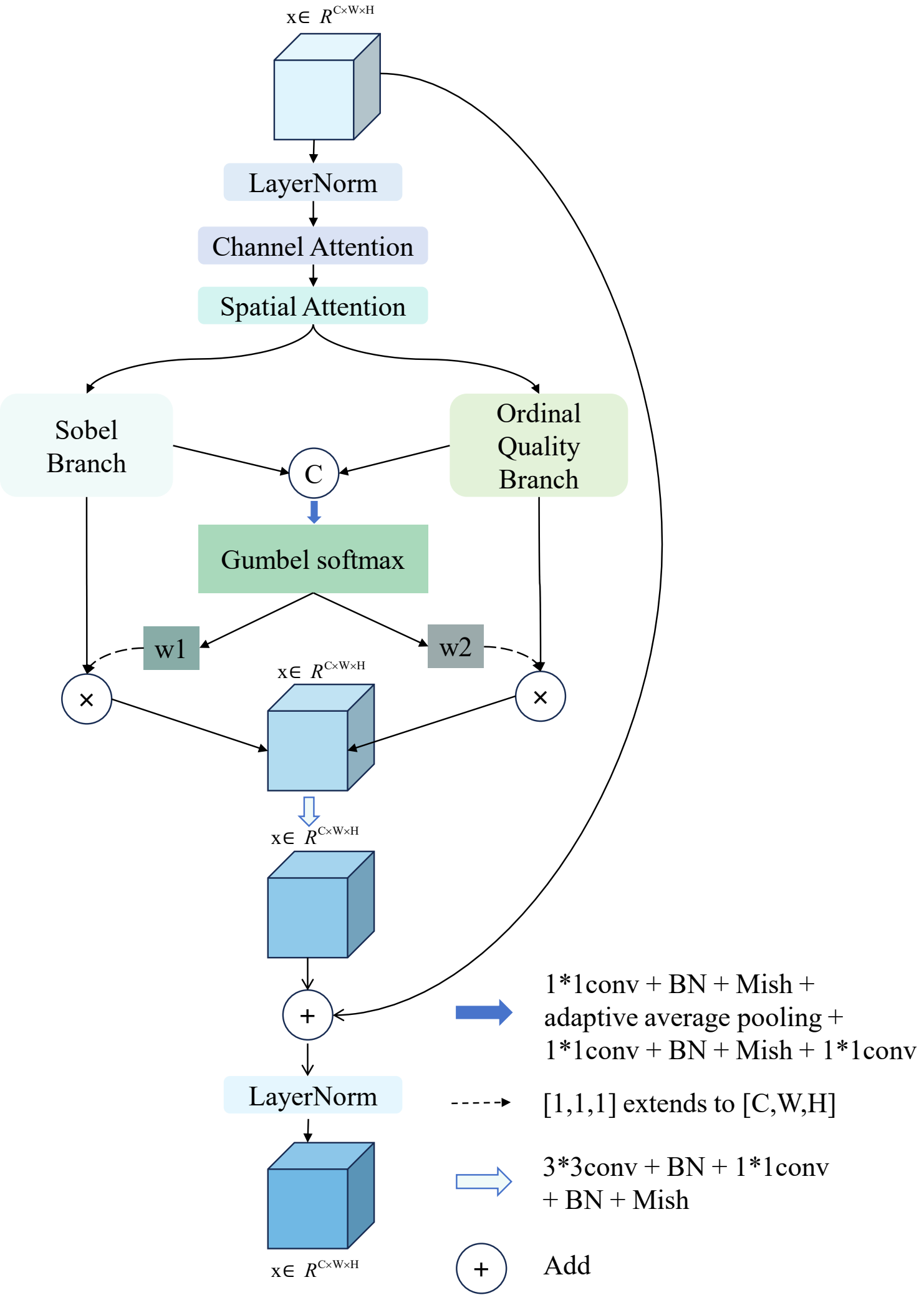

**Fig. 2** The framework of SOQN.

Finally, the fused feature is enhanced by convolution, and then a residual connection is made with the input feature through a learnable weight. The final output is normalized again to obtain the output feature of the SOQN module. This can be expressed by Equation (2) and Equation (3).

$$F_{enh} = f_{enh}(F_{fused}) \tag{2}$$

$$Y = \gamma \cdot F_{enh} + \beta \cdot X \tag{3}$$

where $f_{enh}(\cdot)$ represents the final feature enhancement module, which includes 3*3convolution, batch normalization, 1*1convolution, batch normalization and activation function Mish, $F_{enh}$ represents the enhanced feature, $\gamma$ and $\beta$ represent learnable residual connection weights, $Y$ represents the feature after residual fusion, and $X_{output}$ represents the final output feature of the SOQN module.

The SOQN module adopts a dual-branch structure, including the Sobel branch and the ordinal quality branch.

The Sobel branch is shown in **Fig. 3**, and it realizes the efficient utilization of image boundary information through the following process: The Sobel branch first uses Sobel kernels in four directions (horizontal, vertical, main diagonal, and secondary diagonal) to perform grouped convolution on the input feature map, and comprehensively captures edge details of the image in each main direction through multi-directional edge feature extraction. To further enhance the expression of key edges, the Sobel feature in each direction is input into an independent lightweight sub-network. The structure of this sub-network includes 1×1 convolution, batch normalization, activation function Mish, 1×1 convolution, batch normalization, and activation function Mish, which can dynamically generate an adaptive weight map based on the input Sobel feature. Multiplying this weight map element-wise with the Sobel feature in the corresponding direction allows the network to automatically highlight important edge regions and suppress useless information, thereby improving the discriminative ability of edge features.

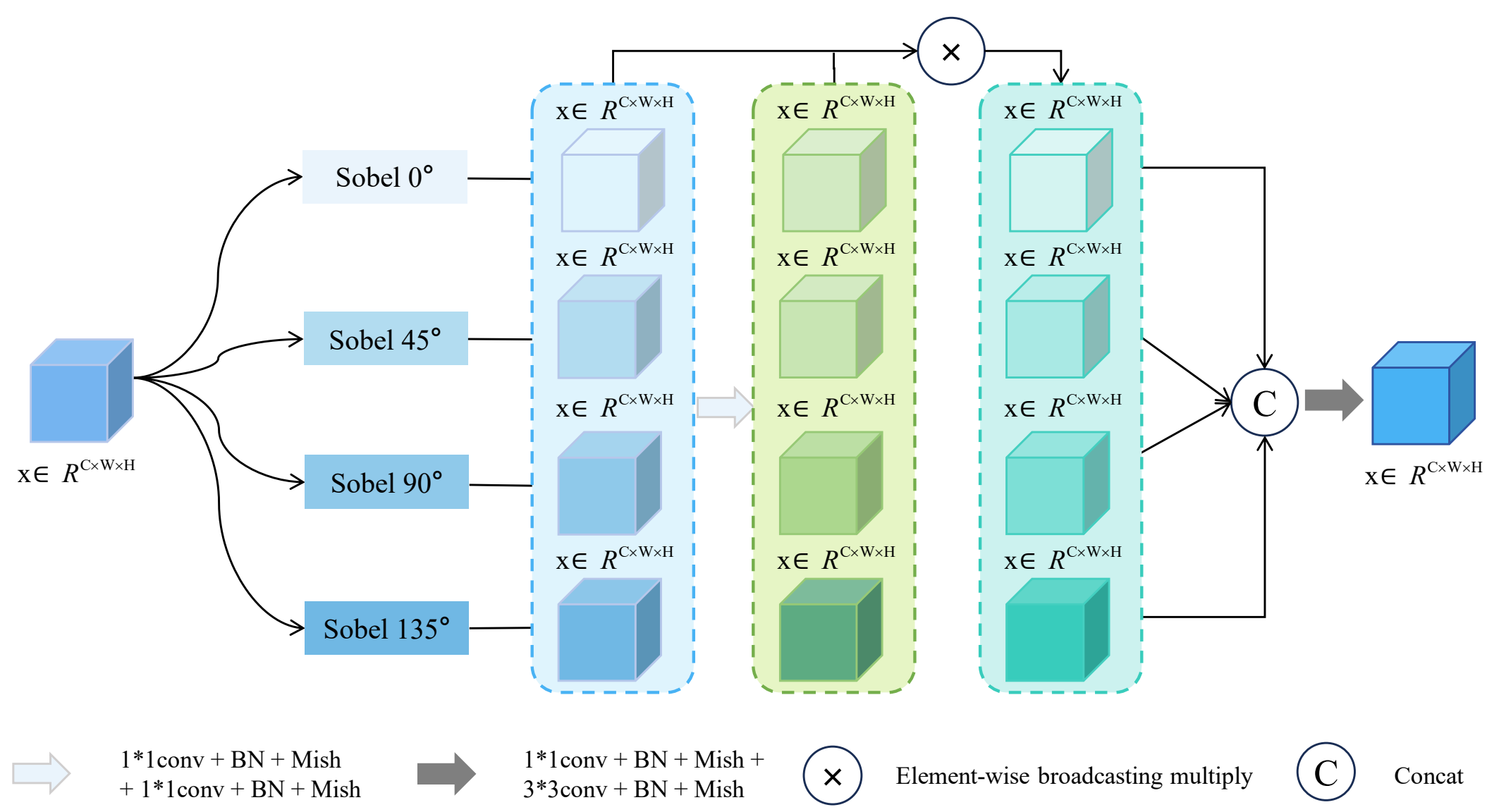

**Fig. 3** The framework of Sobel branch.

Subsequently, the weighted Sobel features from all directions are concatenated in the channel dimension to form a multi-dimensional edge feature set. To fully fuse multi-directional information and compress redundancy, the concatenated features are processed by a convolution fusion network1: first, 1×1 convolution is used to complete dimensionality-increasing integration. After passing through the batch normalization and activation function Mish, 3×3 convolution is used to reduce the number of channels back to the original and deepen the fusion of spatial information. Finally, after passing through convolution fusion network2, which including 1x1 convolution, batch normalization operation, activation function Mish, 3x3 convolution (with padding=1), batch normalization operation, and activation function Mish, a comprehensive representation containing rich edge information is output.

To summarize, the calculation process of the Sobel branch is as follows:

$$S_{out} = g(c(\{h(X,K_i) \odot f(h(X,K_i))\}_{i=1}^{4})),\ i=1,2,3,4 \quad (4)$$

where $X$ denotes the input feature, $K_i$ denotes the $i$-th Sobel kernel, $h(X,K_i)$ represents the convolution operation performed on input $X$ with convolution kernel $K_i$, $f$ is the convolution fusion network1, $\odot$ represents element-wise multiplication, $c(\cdot)$ represents concatenation in the channel dimension, $g$ is the convolution fusion network, and $S_{out}$ is the finally output fused edge feature.

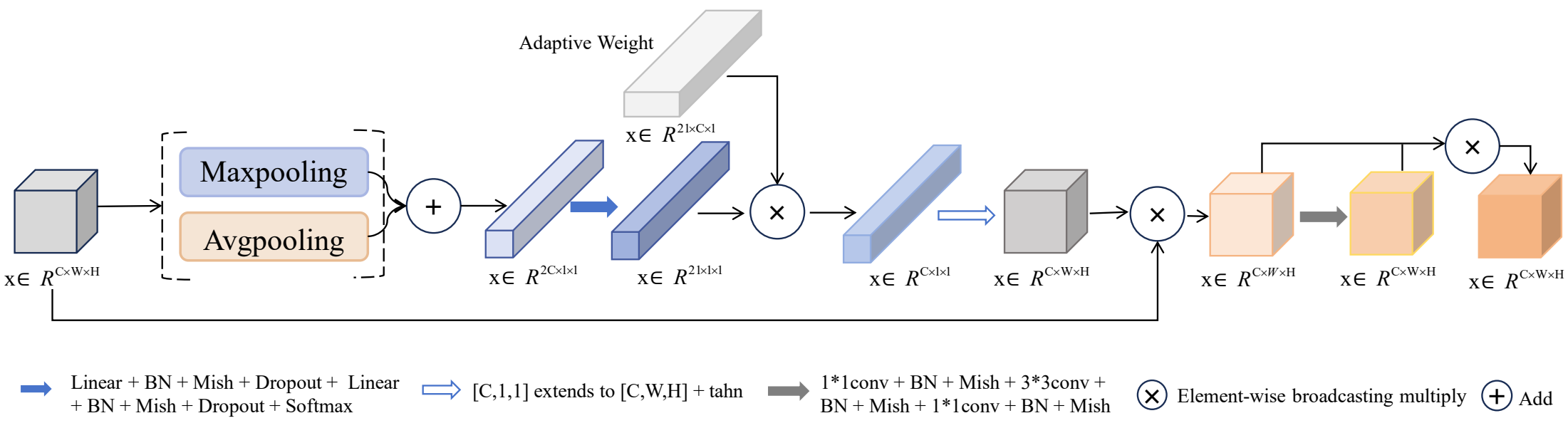

**Fig. 4** The framework of ordinal quality branch.

The ordinal quality branch effectively leverages the ordered nature of labels, and its structure is shown in **Fig. 4**. The specific process is as follows: First, this branch performs global average pooling and global max pooling on the input feature map, then concatenates the two results along the channel dimension to obtain multi-scale global feature representation. Subsequently, this concatenated feature is processed through a multi-layer fully connected network, which consists of three Linear layers. Each layer is followed by batch normalization, Mish activation function, and Dropout in sequence. The last layer outputs scores for each category, which are then processed by the softmax function to obtain the quality probability distribution. Based on this probability distribution, it is multiplied with a learnable ordinal weight matrix to generate adaptive channel weights. After normalization by LayerNorm, these weights are mapped back to the spatial dimension and element-wise multiplied with the preprocessed feature map, achieving ordinal-guided feature reweighting. Finally, the reweighted features are further processed through the convolution enhancement network: first, 1×1 convolution is used for dimension reduction, followed by batch normalization and activation; then 3×3 group convolution is applied to extract spatial features, followed by another batch normalization and activation; finally, 1×1 convolution is used to restore the original number of channels, and after batch normalization and Mish activation, the final output of the ordinal branch is obtained. Throughout the process, this branch realizes adaptive feature modulation based on global quality awareness, enhancing the model's sensitivity and discriminative ability for regions of different quality levels.

In summary, the calculation process of the Ordinal Quality Branch is as follows:

$$O_{out} = (\mathrm{X} \odot \mathrm{t}(\mathrm{L}(\sigma_2(h(c(X_{GAP}, X_{GMP})) * W_{ord})))) \odot e(X_{ord}) \tag{5}$$

where $X$ denotes the input feature map, $X_{GAP}$ and $X_{GMP}$ denote the feature maps after global average pooling and global max pooling respectively, $c(\cdot)$ denotes concatenation along the channel dimension, $h(\cdot)$ denotes the multi-layer fully connected network, $W_{ord}$ denotes the learnable ordinal weight, $L(\cdot)$ denotes channel normalization, $w_{ch}$ denotes the normalized channel weight, $t(\cdot)$ denotes the tanh activation, $*$ denotes matrix multiplication, $e(\cdot)$ denotes the convolution enhancement network, $X_{ord}$ denotes the reweighted feature, and $O_{out}$ denotes the final output of the branch.

### 3.3. Window (W-MTMSA) and shifted window (SW-MTMSA) multi-scale temperature multi-head self-attention.

The (S)W-MTMSA modules is window-based multi-head self-attention mechanism that combines multi-stage attention and quality-aware modulation, aiming to enhance the multi-scale representation capability of features. The structure of (S)W-MTMSA is shown in **Fig. 5**, and the main process is as follows:

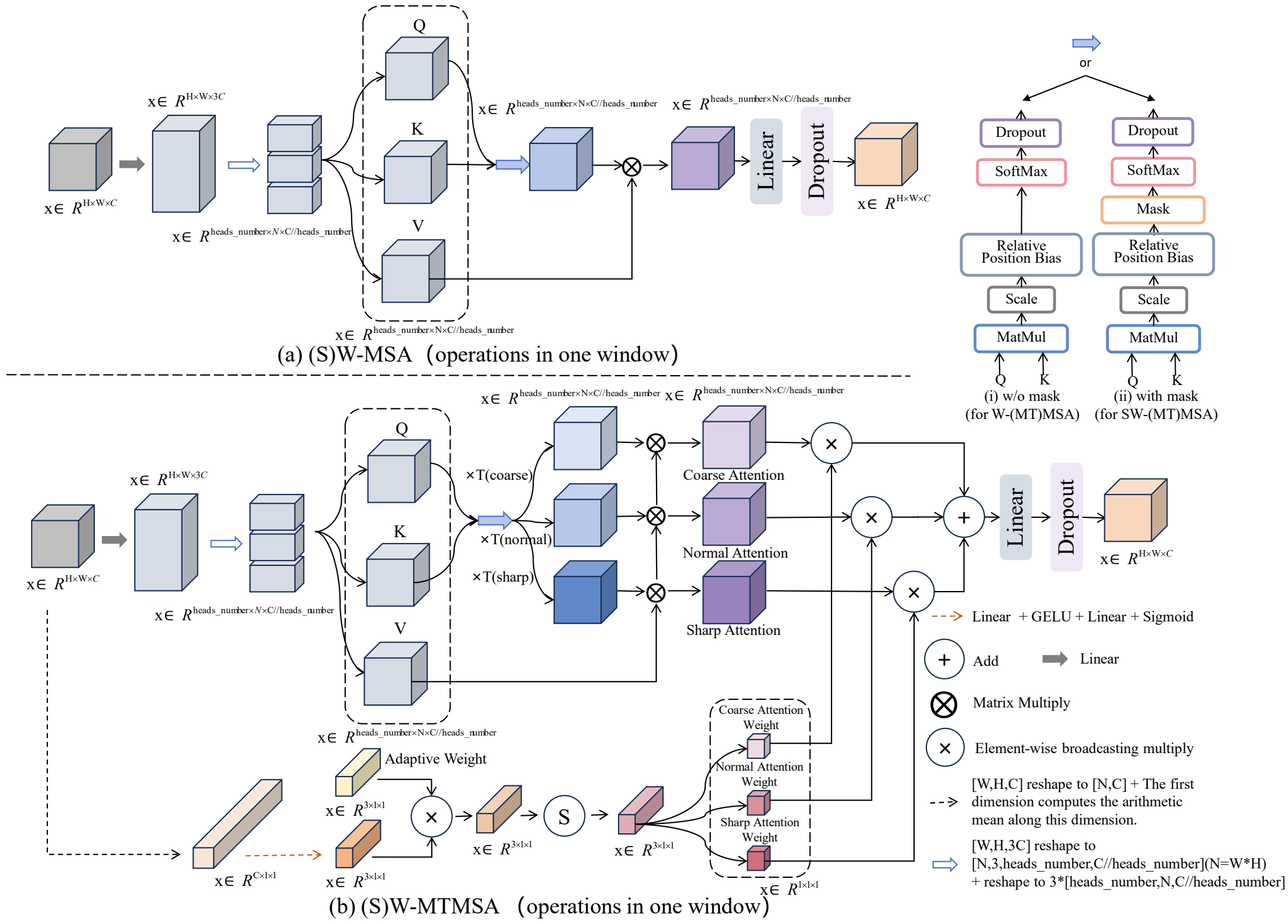

**Fig. 5** Comparison of our proposed (S)W-MTMSA with the traditional (S)W-MSA in the Swin transformer block. (a) The architecture of (S)W-MSA; (b) The architecture of (S)W-MTMSA.

The SW-MTMSA module is a window based multi-head self-attention mechanism designed specifically for multi sharpness feature extraction and quality perception fusion. The detailed process is as follows: the input feature map is first filled to ensure that the height and width dimensions can be evenly divided by the window size, and then decomposed into several independent windows through window segmentation operations, and attention calculation is independently performed within each window. For the features in each window, query (Q), key (K), and value (V) vectors are generated through linear transformation:

$$Z = X * W_{qkv} + b_{qkv} \tag{6}$$

$$Z' = p(r(Z,(B,N,3,h,d)),(2,0,3,1,4)) \tag{7}$$

$$[Q,K,V] = [Z'_0, Z'_1, Z'_2] \tag{8}$$

where $X$ is the input feature within the window, $W_{qkv}$ is the learnable weight, $b_{qkv}$ is the learnable bias, $p(\cdot,\cdot)$ represents permute, $r(\cdot,\cdot)$ represents reshape, B, H, W, and C respectively represent the batch size, height, width, and number of channels of the feature map, N is the product of H and W, h is the number of attention heads, and $d$ is the number of channels in each classification head.

Meanwhile, a learnable relative position bias is introduced into the attention score calculation, and the bias table is invoked through preset indices to enhance the capability of spatial relationship modeling:

$$A = \frac{Q*K^T}{\sqrt{d}} + b_{rel} \tag{9}$$

where $b_{rel}$ represents the relative position bias.

For the SW-MTMSA module, a mask mechanism needs to be introduced to achieve occlusion between windows. The specific operation is to add $-\infty$ to the positions where attention interaction is not allowed. After the softmax operation, the attention weights at these positions will be set to 0, thereby completing the occlusion between windows. Finally, perform Dropout. The process can be described by the following formulas:

$$A' = \mathrm{r}((r(A, (B/nW,\ nW,\ h,\ N,\ N)) + M), (-1,\ \mathrm{h},\ \mathrm{N},\ \mathrm{N})) \tag{10}$$

$$Attn' = D(\sigma_2(A')) \tag{11}$$

where $M$ represents the mask, taking $-\infty$ in positions where attention interaction is not allowed, and 0 in other places. $nW$ denotes the number of the windows. $D(\cdot)$ denotes dropout. SW-MTMSA implements the calculation of attention distributions with three temperature parameter adjustments in parallel within a single window: the coarse stage uses a lower temperature parameter to make the probability distribution smoother to capture global rough correlations; the standard stage uses a conventional temperature parameter to generate a balanced attention distribution; the sharpening stage enhances the sharpness of the probability distribution through a higher temperature parameter to highlight local salient features. The three stages are fused with the value vector ($V$) respectively to obtain three sets of feature outputs. For the SW-MTMSA module, the specific formula is:

$$O_i = \sigma_2((Attn' * T_i) * \mathrm{V})\ i = low, medium, high \tag{12}$$

where $T_{low}$ , $T_{medium}$ and $T_{high}$ represent values for low temperature, medium temperature, and high temperature, respectively, while $O_{low}$ , $O_{medium}$ , and $O_{high}$ represent attention at low temperature, medium temperature, and high temperature, respectively.

The module has a built-in quality-aware modulator. The mean value $\bar{X}$ of the features within the window is processed through a two-layer fully connected network and an activation function, outputting three modulation weights corresponding to the three stages, activated by Sigmoid:

$$q = \rho(W_2 * \phi(W_1 * \bar{X})) \tag{13}$$

where $\phi$ denotes the activation function Mish, $\rho$ denotes sigmoid, and $W_1$ and $W_2$ are fully connected weights.

At the same time, learnable progressive weights $w_i^{proj}$ is introduced to model the global importance of the three stages. Finally, the output features of the three stages are normalized by softmax through the product of quality-aware weights and progressive weights to obtain fusion coefficients:

$$\alpha_i = \sigma_2(q \odot \mathrm{S}(w_i^{proj}))\ i = low, medium, high \tag{14}$$

And perform weighted summation on the three sets of features:

$$O_{final} = \sum_{i \in \{low, medium, high\}} \alpha_i\, O_i \tag{15}$$

The fused features undergo linear mapping and dropout processing to obtain the window-level output:

$$Y = D(O_{final} * W_O) \tag{16}$$

where $D(\cdot)$ denotes dropout.

All window outputs are restored to the original spatial structure through an inverse partitioning operation and the initial padding part is removed.

For the W-MTMSA module, no masking mechanism, window shift, and reverse window shift are introduced, and all other steps are consistent with the SW-MTMSA module, just like the difference between W-MSA and SW-MSA.

### 3.4. Hierarchical ranked probability score (HRPS) loss

Rank probability score (RPS) [30] is an appropriate scoring rule suitable for ordered classification,

which can impose more significant penalties on predictions that are far from the true category. The agaldran team transformed RPS into RPS loss to adapt to medical image grading tasks [16]. Inspired by the agaldran team's RPS loss [16], we designed the HRPS loss. The HRPS loss module is a custom loss function specifically designed for multi-level classification tasks. It can balance the ranking consistency and distribution diversity of fine (21 classes) and coarse (7 classes) labels. Its implementation process is as follows: In the initialization phase, the module establishes the corresponding relationship between fine categories and coarse categories through a preset mapping table, and records the indices of fine categories included in each coarse category. This provides efficient lookup support for subsequent probability aggregation and label conversion.

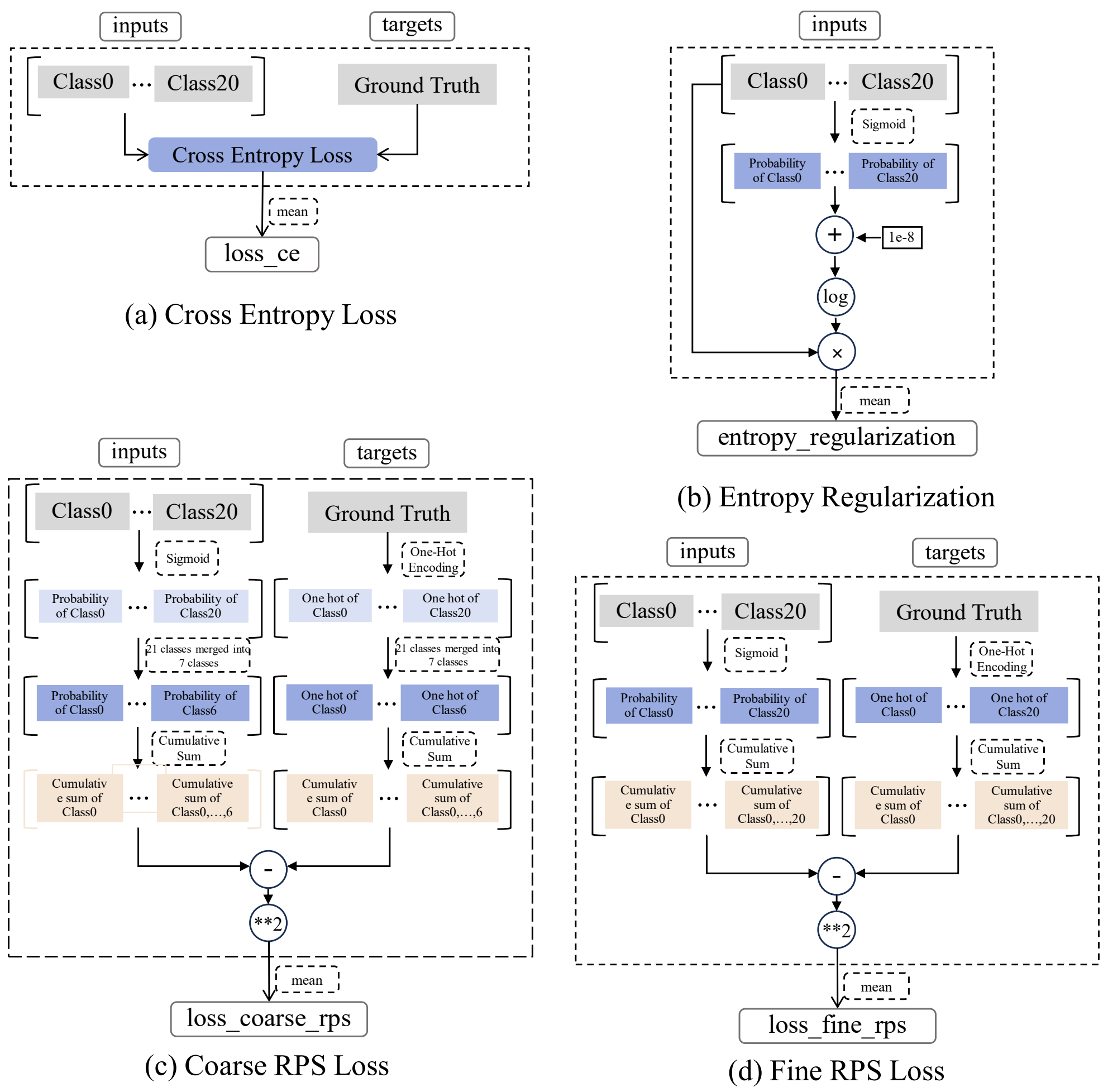

**Fig. 6** The calculations of components of HRPS Loss

During forward propagation, first, the ground-truth labels are one-hot encoded. The predicted features are processed by the softmax function to obtain a probability distribution. The standard cross-entropy loss is calculated for the two, serving as the basic classification loss, and its structure is shown in **Fig. 6a**:

$$L_{CE} = -\sum_{i=1}^{C} y_i \, log\hat{y}_i \tag{17}$$

where $C$ is the number of fine categories, $\hat{y}_i$ is the predicted probability of the $i$-th category after softmax, and $y_i$ represents the one-hot encoding of the $i$-th category. Subsequently, the predicted features are processed by the softmax function again to obtain a probability distribution. The cumulative sums are calculated for the one-hot encoded labels and the probability distribution of the predicted features respectively:

$$y_i^{cumsum} = \sum_{j=1}^{i} y_j, i = 0,1,\ldots,C \tag{18}$$

where $y_i^{cumsum}$ represents the cumulative sum of the one-hot encoding of the *i*-th category.

$$\hat{y}_i^{cumsum} = \sum_{j=1}^{i} \hat{y}_j, i = 0,1,\ldots,C \tag{19}$$

where $\hat{y}_i^{cumsum}$ represents the cumulative sum of the probability distribution of the predicted features for the *i*-th category, and $\hat{y}_i$ represents the probability distribution of the predicted features for the *i*-th category. Then, the difference of the squares between the cumulative sum of the one-hot encoded labels and the cumulative sum of the probability distribution of the predicted features is calculated for each label, obtaining the fine RPS loss, which is used to measure the consistency between the model's output probability distribution and the ordering of the ground-truth labels, and its structure is shown in **Fig. 6d**:

$$L_{HRPS}^{fine} = \sum_{i=0}^{C} (\hat{y}_i^{cumsum} - y_i^{cumsum})^2 \tag{20}$$

Next, a preset mapping relationship is used to aggregate fine labels and probabilities into coarse categories: The fine labels and their probabilities are aggregated into coarse categories through a preset mapping relationship: three consecutive categories are merged into one large category, that is, 21 categories are mapped to 7 categories. Let $\varphi: \{0,1,\ldots,C\} \rightarrow \{0,1,\ldots,K\}$ denote the mapping function from fine categories to coarse categories, where K is the number of coarse categories. Then, the coarse labels are also one-hot encoded, and the probability distribution is aggregated:

$$z_i = \begin{cases} 1, & if\ there\ exists\ k \in \varphi^{-1}(i)\ such\ that\ y_k = 1 \\ 0, & otherwise \end{cases} \tag{21}$$

where $\varphi^{-1}(i)$ represents all fine categories mapped to the *i*-th coarse category, and $z_i$ is the one-hot code of the *i*-th coarse category.

$$\hat{z}_i = \sum_{k \in \varphi^{-1}(i)} \hat{y}_k \tag{22}$$

where $\hat{z}_i$ represents the predicted probability distribution of the *i*-th coarse category. Then, the coarse RPS loss is calculated by the square of the difference between the two cumulative sums, further constraining the rationality of the model's ranking at the large category level, and its structure is shown in **Fig. 6c**:

$$z_i^{cumsum} = \sum_{j=1}^{i} z_j, i = 0,1,\ldots,K \tag{23}$$

$$\hat{z}_i^{cumsum} = \sum_{j=1}^{i} \hat{z}_j, i = 0,1,\ldots,K \tag{24}$$

$$L_{HRPS}^{coarse} = \sum_{i=0}^{K} (\hat{z}_i^{cumsum} - z_i^{cumsum})^2 \tag{25}$$

where $z_i^{cumsum}$ and $\hat{z}_i^{cumsum}$ are the cumulative sum of the one-hot codes of the *i*-th coarse category and the predicted feature probability distribution of the one-hot code of the *i*-th category, respectively.

The fine and coarse RPS losses are dynamically adjusted by scaling their gradients to align with the magnitude of the cross-entropy loss gradient, following the MetaBalance [39]. This avoids manual step adjustment issues caused by gradient magnitude discrepancies between different loss terms, ensuring that each loss term contributes appropriately to the parameter updates without being dominated by or overshadowed by others:

$$L_{HRPS,norm}^{fine} = \frac{L_{HRPS}^{fine}}{\frac{\|L_{HRPS}^{fine}\|}{\|L_{CE}\|} + \epsilon} \tag{26}$$

$$L_{HRPS,norm}^{coarse} = \frac{L_{HRPS}^{coarse}}{\frac{\|L_{HRPS}^{coarse}\|}{\|L_{CE}\|} + \epsilon} \tag{27}$$

where $\epsilon$ denotes a small value of 1e-8.

The sum of the three constitutes the main loss output:

$$L_{main} = \lambda_1 L_{CE} + \lambda_2 L_{HRPS,norm}^{fine} + \lambda_3 L_{HRPS,norm}^{coarse} \tag{28}$$

In addition, the module incorporates an entropy regularization mechanism [40] by calculating the information entropy of the predicted probability distribution, and its structure is shown in **Fig. 6b**:

$$H(p) = -\sum_{i=1}^{C} p_i log(p_i) \tag{29}$$

The negative entropy is added to the total loss as a regularization term to make the model's output distribution smoother and more diverse, preventing overconfidence.

The final loss is:

$$L_{total} = L_{main} + \lambda_4 \cdot (-H(p)) \tag{30}$$

Finally, the forward propagation process returns the total loss. The overall structure is concise and easy to tune, which can effectively improve the model's performance in multi-granularity ranking consistency and distribution diversity, and is suitable for fine grading tasks such as medical imaging.

## 4. Experiments and Results

### 4.1. Dataset

LDCTIQAG2023 [16]: The LDCTIQAG2023 dataset is a dedicated evaluation dataset for the 2023 Low-Dose Computed Tomography (LDCT) Grand Challenge, and its core value lies in its high simulation of real clinical scenarios. This dataset contains 1,500 contrast-enhanced abdominal CT images, all with a uniform size of 512×512 pixels. 6 radiologists independently score each image based on a unified standard (the scoring range is 0-4, where 0 represents bad quality, 1 represents poor quality, 2 represents fair quality, 3 represents good quality, and 4 represents excellent quality). The average score of the six radiologists is finally taken as the quality score label of the image. **Fig. 7** displays 8 abdominal CT images with different quality scores. It can be observed that as the score decreases, noise and artifacts in the images become increasingly pronounced.

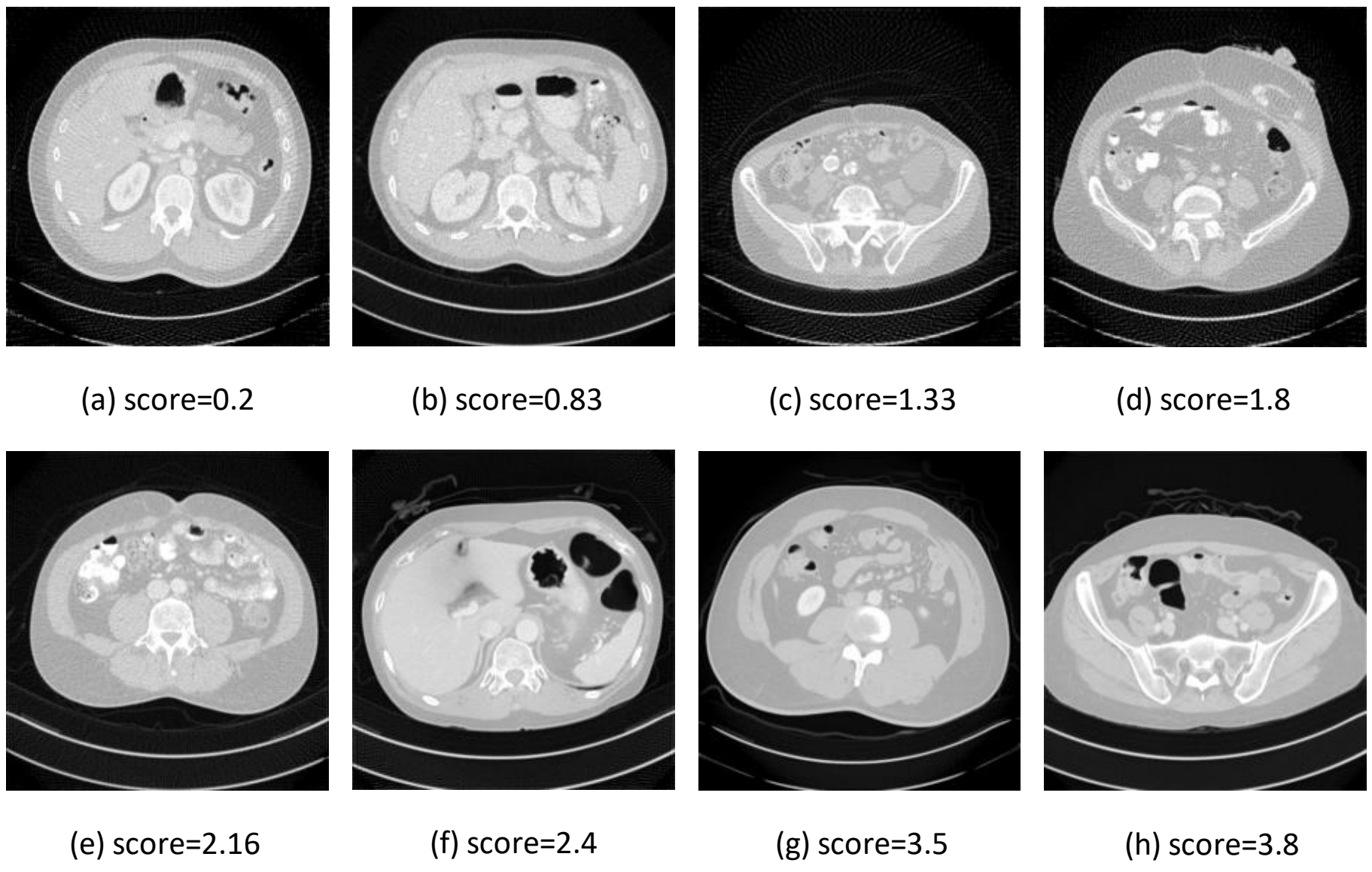

**Fig. 7** The images of LDCTIQAG2023 Dataset.

The original division of the dataset includes a training set (1,000 images), a validation set (200

images), and a test set (300 images), among which the training set[1] and the test set[2] are publicly available. To avoid data leakage, the LDCTIQAG2023 challenge has explicitly ensured that the patients in the training set and test set are completely independent. To meet the model training and validation requirements of this study, we adjusted the original data division: 800 images were split from the original 1,000-image training set as the new training dataset, and the remaining 200 images were used as the new validation dataset; the original 300 images in the test set remained unchanged and were directly used as the test dataset of this study. In addition, to meet the training needs of the model's classification task, we further processed the image quality labels: the original average score (0-4) of each image was multiplied by 5 and then rounded to obtain classification labels ranging from 0 to 20, finally forming 21 discrete classification categories.

### 4.2. Evaluation metrics

To evaluate the effectiveness of the model performance, we select four categories of metrics from the field of image quality assessment (IQA), which are specifically designated for the Low-dose Computed Tomography Perceptual Image Quality Assessment Grand Challenge 2023 [16]. The selection of these metrics is fully consistent with the challenge standards, allowing direct performance comparison with the methods proposed in the challenge. It includes three classic correlation indicators: the Pearson's linear correlation coefficient (PLCC) can accurately capture the linear correlation between the model's predicted results and the radiologists' subjective assessment results; the Spearman's rank-order correlation coefficient (SROCC) and Kendall's rank-order correlation coefficient (KROCC) can effectively reflect the non-linear rank correlation between the two, avoiding the limitations of linear assumptions on evaluation results [41]. Meanwhile, a comprehensive metric—the sum of the absolute values of the three correlation coefficients—is also included [16]. By integrating information from the three basic metrics, this comprehensive indicator quantifies the degree of agreement between the model output and expert assessments more intuitively from an overall perspective. Based on these unified challenge-specified metrics, the performance differences between the proposed model and existing methods in the challenge can be clearly compared, and the advantages and positioning of the proposed model in the low-dose CT perceptual image quality assessment task can be clearly identified.

First, the PLCC was employed to assess the linear correlation, and its calculation is as follows:

$$PLCC = \frac{\sum(\hat{s}_i-\hat{\mu})(s_i-\mu)}{\sqrt{\sum(\hat{s}_i-\hat{\mu})^2(s_i-\mu)^2}}, \tag{31}$$

where $\hat{s}_i$ denotes the i-th predicted image quality score, and $s_i$ signifies the i-th averaged score furnished by radiologists. Correspondingly,$\hat{\mu}$ and $\mu$ are respective of the mean values for the predicted scores and the scores from radiologists.

In addition, SROCC and KROCC are used to evaluate nonlinear associations. The calculation formula for SROCC is as follows:

$$SROCC = 1 - \frac{6\sum(\widehat{S}_i-S_i)^2}{n(n^2-1)}, \tag{32}$$

where $\widehat{S}_i$ and $S_i$ denote the ranks corresponding to the i-th element in the predicted image quality score set S and the radiologists' assessment score set S, respectively. The variable n signifies the total count of images.

[1] https://zenodo.org/records/7833096#.ZEFywOxBzn5

[2] https://drive.google.com/file/d/1DbZfTj4it-3QRTuJO73r1iQyHLRwPS6H/view

Additionally, the calculation of KROCC is performed as follows:

$$KROCC = \frac{M_c - M_d}{\sqrt{(M_c + M_d + T_{\hat{S}})(M_c + M_d + T_S)}}, \tag{33}$$

where $M_c$ and $M_d$ correspond to the quantities of concordant pairs and discordant pairs, respectively. Additionally, $T_{\hat{s}}$ and $T_s$ indicate the counts of ties that are exclusively present within the predicted score set $\hat{S}$ and the radiologists' score set S, respectively. The ultimate overall correlation score was derived by integrating the values of all correlation coefficients:

$$Score = |PLCC| + |SROCC| + |KROCC|, \tag{34}$$

### 4.3. Implementation details

The experiments are conducted on an NVIDIA RTX 4090 Graphics Processing Unit (GPU). Before feeding the images into the model, all images undergo normalization to ensure training stability. To avoid overfitting and enhance the model's generalization ability, this study employs basic data augmentation techniques provided in the PyTorch framework (see **Tab. 1** for details). The Nadam optimizer is selected, along with a cosine annealing learning rate scheduler with cyclic restarts. The initial learning rate is set to 1e-4, and the minimum learning rate is 0. For the LDCTIQAG2023 dataset, the batch size is set to 8, and the model is trained for a total of 80 epochs. In the SOQN model, the initial values of the adaptive weights $\beta$ and $\gamma$ for weighted residual connections are 0.8 and 0.2 When adopting (S)W-MTMSA, the initial values of the adaptive parameters $T_{low}$, $T_{medium}$, and $T_{high}$ are set to 0.5, 1.0, and 2.0 respectively, and the initial values of three different sharpness attention adaptive fusion weights $w_{low}^{proj}$, $w_{medium}^{proj}$, and $w_{high}^{proj}$ are 0.3, 0.4, and 0.3. In the case of the hierarchical ranked probability score (HRPS) loss function, the parameters $\lambda_1$, $\lambda_2$, and $\lambda_3$ were all assigned a value of 1, and the parameter $\lambda_4$ was set to 5e-5.

**Tab. 1** Data augmentation in our study.

| Data Augmentation | values |
|---|---|
| RandomRotation | ±15° |
| RandomAffine | (0.95,1.20) |
| RandomHorizontalFlip | True |
| RandomVerticalFlip | True |

### 4.4. The performance of ClinReadNet

In this section, performance comparison experiments are conducted on the LDCTIQAG2023 dataset. The evaluation objects of the proposed model are divided into two categories of methods: one is the commonly used general deep convolutional neural network methods at the current stage, and the other is other advanced models specially developed for the abdominal Computed Tomography (CT) image quality assessment task.

The agardran model is based on Swin Transformer [26] and BiT ResNeXt-50 [27], and adopts a multi backbone network, multi loss function, and multi-head ensemble architecture [28]. Each head is supervised by cross entropy loss for training. After taking the average of the prediction results for each head, the ranking probability loss [29, 30] was used to minimize the ranking loss, achieving higher accuracy in distinguishing different image quality levels and improving prediction stability. The RPLAXIS model adopts a multi-dimensional attention network [31] and scale aware design, enabling it to accurately quantify image quality without relying on reference images, making it suitable for no reference assessment scenarios. CHILL@UK has improved the input and output layers of the lightweight backbone network EfficientNetV2-t [42] and adopted a model integration strategy [16]. This strategy improves the accuracy of image quality regression while maintaining good computational efficiency. FeatureNet integrates a multi-dimensional feature extraction module, which can comprehensively capture micro details and structural information related to image quality, thereby improving sensitivity to quality degradation [16]. The model proposed by Team Epoch adopts a two-stage multi-task (regression and classification) framework, which focuses on noise features in stages and can accurately characterize the quality degradation of low-dose computed tomography (LDCT) images [16]. The Gabybaldeon utilizes knowledge distillation techniques: in the initial stage, a high-performance teacher model is trained, and in the subsequent stage, knowledge is transferred to a lightweight student model to achieve a balance between performance and inference efficiency [16].

For the completeness of the comparative experiments, this study also includes the convolution-based architectures MobileNet_v2 [43], ResNet50 [44], ConvNeXt [45], VGG16 [46], and the Transformer-based architectures BEiT [47], Vision Transformer (ViT) [32], and Swin-T [26] into the evaluation scope. These models have fully demonstrated their excellent performance in various visual tasks in the past and can provide an effective reference for the performance of general deep learning models.

Regarding the approaches documented in Ref. [16], since the datasets, data partitioning methods, and task configurations adopted in these studies are consistent with those of this study, direct comparison is considered valid; based on this, this study directly adopts the experimental results of these methods published in their academic papers. As for other methods, they are re-implemented under the experimental conditions set in this study.

**Tab. 2** presents the performance comparison between our method and various models on the LDCTIQAG2023 dataset, covering the state-of-the-art agaldran model, commonly used general deep convolutional neural network methods at the current stage, and other advanced models specifically developed for the abdominal Computed Tomography (CT) image quality assessment task. In the performance evaluation, we use Score as the core comprehensive metric, and specifically, it is the sum of the absolute values of the three metrics: PLCC, SROCC, and KROCC. Our method achieves a Score of 2.7690 in this metric, which is significantly superior to all other comparative models listed in the table.

Among the models participating in the comparison, the agaldran model performs the second best with a Score of 2.7427, followed by the ConvNeXt model with a Score of 2.7213. The Scores of the RPI_AXIS model and ResNet50 model are 2.6843 and 2.6802, respectively; the Scores of the MobileNet_v2 model and CHILL@UK model are 2.6719 and 2.6550 in sequence; the Team Epoch model has a Score of 2.6202; while the VGG16 model has a Score of 0.7253, and it should be noted that this value shows a significant gap compared with other models. Furthermore, the gabybaldeon model has the lowest Score of only 2.5671, and its values in the three individual metrics (PLCC, SROCC, KROCC) are also lower than those of other comparative models, resulting in the weakest overall performance.

In terms of the three key individual metrics (PLCC, SROCC, KROCC), our method also demonstrates significant advantages, reaching 0.9507, 0.9554, and 0.8629 respectively—all higher than the corresponding metric values of all other comparative models. This result fully confirms that our method exhibits good robustness under multi-dimensional evaluation metrics, with stable and reliable performance. In summary, the comparison results clearly indicate that our model outperforms various other advanced models both in the comprehensive metric "Score" and in individual performance metrics such as PLCC, SROCC, and KROCC, fully demonstrating the superiority of our method in the abdominal CT image quality assessment task.

**Tab. 2** Comparison with the existing state-of-the-art (SOTA) methods.

| Method | | PLCC | SROCC | KROCC | Score |
|---|---|---|---|---|---|
| General classification methods | Mobilenet_v2 [43] | 0.9284 | 0.9318 | 0.8122 | 2.6724 |
| | Resnet50 [44] | 0.9184 | 0.9362 | 0.8255 | 2.6802 |
| | BeIT [47] | 0.9381 | 0.9400 | 0.8261 | 2.7042 |
| | Vision Transformer [32] | 0.9382 | 0.9446 | 0.8377 | 2.7204 |
| | ConvNeXt [45] | 0.9402 | 0.9449 | 0.8361 | 2.7213 |
| | Swin-T [26] | 0.9414 | 0.9452 | 0.8373 | 2.7239 |
| | VGG16 [46] | 0.9415 | 0.9461 | 0.8390 | 2.7265 |
| SOTA methods in LDCT grand challenge | gabybaldeon [16] | 0.9143 | 0.9096 | 0.7432 | 2.5671 |
| | Team Epoch [16] | 0.9278 | 0.9232 | 0.7691 | 2.6202 |
| | FeatureNet [16] | 0.9362 | 0.9338 | 0.7851 | 2.6550 |
| | CHILL@UK [16] | 0.9402 | 0.9387 | 0.7930 | 2.6719 |
| | RPI_AXIS [16] | 0.9434 | 0.9414 | 0.7995 | 2.6843 |
| | agaldran [16] | 0.9491 | 0.9495 | 0.8440 | 2.7427 |
| | **ClinReadNet (ours)** | **0.9507** | **0.9554** | **0.8629** | **2.7690** |

### 4.5. Ablation study

To evaluate the independent contribution of each component in the proposed model, we conduct ablation experiments on the LDCTIQAG2023 dataset. As shown in **Tab. 3**, the experiment uses Swin-T [26] as the baseline model. By gradually integrating different core components and iteratively optimizing the network structure, the complete ClinReadNet network is finally constructed. The specific improvement process and performance results of each stage are as follows:

Firstly, we introduced the SOQN module into the model and embed it between the first and second stages of Swin-T to enhance the ability to retain key information during feature transmission. After this structural modification, the model's Score reaches 2.7419, which is 0.018 higher than that of the baseline model. This demonstrates the positive contribution of the SOQN module on model performance.

**Tab. 3** Ablation study of the proposed framework.

| Method | PLCC | SROCC | KROCC | Score |
|---|---|---|---|---|
| Swin-T | 0.9414 | 0.9452 | 0.8373 | 2.7239 |
| Swin-T+SOAN | 0.9440 | 0.9507 | 0.8471 | 2.7419 |
| Swin-T+SOAN+(S)W-MTMSA | 0.9491 | 0.9539 | 0.8582 | 2.7611 |
| **ClinReadNet (Ours)** | **0.9507** | **0.9554** | **0.8629** | **2.7690** |

Secondly, we replaced the W-MSA and SW-MSA of the SwinTransformer Block in stage 4 of Swin-T with W-MTMSA and SW-MTMSA, respectively. After this module replacement, the model score further increased by 0.0192 to 2.7611, indicating that the (S)W-MTMSA module has a significant effect on enhancing feature representation ability.

Finally, to further optimize the loss calculation logic during the model training process, we introduce the HRPS loss function. This final enhancement specifically refers to the construction of the full ClinReadNet architecture, which allows the model's Score to rise by an additional 0.0089 and eventually attain a value of 2.7690. This becomes the solution with the highest Score among all experimental network configurations, fully demonstrating the comprehensive improvement effect of the synergy between components on model performance.

#### *4.5.1. SOQN module*

The SOQN module consists of two core branches: one is the Sobel branch, which mainly extracts local edge features of images through the Sobel operator and focuses on quality-related information at the image detail level; the other is the ordinal quality branch, which focuses on capturing quality grade-related features across the global scope of images and emphasizes the representation of overall quality trends. To further clarify the functional role of each branch within the SOQN module and their impacts on model performance, we conducted supplementary experiments. The experimental design and operation details are as follows:

In Experiment 1# (SOQN w/o Sobel branch), we removed the Sobel branch from the SOQN module and retained only the ordinal quality branch. In this case, the focus of the SOQN module was solely on the global quality information of images; in Experiment 2# (SOQN w/o ordinal quality branch), we conversely removed the ordinal quality branch from the SOQN module and retained only the Sobel branch. At this point, the SOQN module focused exclusively on the local edge information of images.

**Tab. 4** Ablation study of SOQN module.

| Method | PLCC | SROCC | KROCC | Score |
|---|---|---|---|---|
| 1# SOQN w/o Sobel branch | 0.9468 | 0.9479 | 0.8496 | 2.7443 |
| 2# SOQN w/o ordinal quality branch | 0.9472 | 0.9508 | 0.8523 | 2.7503 |
| **3# ClinReadNet (ours)** | **0.9507** | **0.9554** | **0.8629** | **2.7690** |

As shown in **Tab. 4**, when only the Sobel branch is used, the model achieves a Score of 2.7503; when only the ordinal quality branch is adopted, the Score is 2.7443. Both sets of results are lower than the Score of 2.7690 achieved when both branches are enabled simultaneously.

In the SOQN module, we used adaptive weights for weighted residual connections, as shown in formula (12). Here, we compare the results of the weighted residual connection method with the effects of ordinary residual connections and the effects generated by different initial values of adaptive weights.

**Tab. 5** Ablation study of SOQN module.

| Method | PLCC | SROCC | KROCC | Score |
|---|---|---|---|---|
| 1# SOQN with common residual connection | 0.9479 | 0.9524 | 0.8544 | 2.7548 |
| **2# SOQN with weighted residual connection, $\beta$=0.8 and $\gamma$=0.2 (ours)** | **0.9507** | **0.9554** | **0.8629** | **2.7690** |

As shown in **Tab. 5**, when using the classical residual connection, the Score is only 2.7548, while when using the weighted residual connection, the Score reaches 2.7690. The effect of using the weighted residual connection is better than using the classical residual connection number.

#### *4.5.2. (S)W-MTMSA module*

To further explore the mechanism of action and practical impact of single and combined sharpness attention of (S)W-MTMSA under different temperature scenarios, we designed and conducted additional supplementary experiments.

**Tab. 6** Ablation study of (S)W-MTMSA module.

| Method | PLCC | SROCC | KROCC | Score |
|---|---|---|---|---|
| 1# (S)W-MTMSA w/o low and medium temperature | 0.9463 | 0.9463 | 0.8410 | 2.7336 |
| 2# (S)W-MTMSA w/o low and high temperature | 0.9435 | 0.9479 | 0.8447 | 2.7361 |
| 3# (S)W-MTMSA w/o medium and high temperature | 0.9470 | 0.9480 | 0.8459 | 2.7409 |
| 4# (S)W-MTMSA w/o high temperature | 0.9458 | 0.9492 | 0.8488 | 2.7438 |
| 5# (S)W-MTMSA w/o medium temperature | 0.9460 | 0.9493 | 0.8478 | 2.7431 |
| 6# (S)W-MTMSA w/o low temperature | 0.9458 | 0.9495 | 0.8488 | 2.7442 |
| **7#** (S)W-MTMSA **(Ours)** | **0.9507** | **0,9554** | **0.8629** | **2.7690** |

The experimental design and operation details are as follows: A total of 6 groups of single/combined verification schemes (Experiments 1–6#) were set up in the experiments, and each group used sharpness attention under a specific temperature for testing: Experiments 1–3# were single-temperature sharpness schemes, using independent sharpness attention under low temperature, medium temperature, and high temperature in sequence; Experiments 4–6# were dual/triple-temperature sharpness combination schemes, adopting sharpness attention combinations of "low and medium temperature", "low and high temperature", and "medium and high temperature" respectively; Experiment 7# was a full combination scheme of "low, medium and high temperature", so as to fully cover the possible configurations of sharpness attention under different temperature scenarios.

As shown in **Tab. 6**, the performance differences of the (S)W-MTMSA module under different temperature-sharpness combinations can be clearly summarized: In the full combination scheme of "low, medium and high temperature", the model exhibits the optimal performance with a Score of 2.760; followed by the "medium and high temperature" combination, which achieves a Score of 2.7442; next are the "low and medium temperature" combination and "low and high temperature" combination, with Scores of 2.7438 and 2.7431 respectively. These two schemes perform slightly worse than the previous two combinations but still outperform the single-temperature schemes. Among the single-temperature schemes, the "low temperature" single-sharpness scheme shows relatively prominent performance with a Score of 2.7409; the "medium temperature" single scheme ranks second with a Score of 2.7361, while the "high temperature" single scheme performs relatively weakly among all schemes, also with a Score of 2.7361. Overall, as the coverage range of temperature-sharpness combinations narrows, the model's Score shows a gradual downward trend. This pattern further confirms the effectiveness of the (S)W-MTMSA module in improving performance through multi-temperature domain feature fusion.

To further verify the practical effectiveness of the proposed (S)W-MTMSA in this study, we designed comparative experiments for verification: the original (S)W-MTMSA module in the model was replaced with two classic attention mechanisms—spatial attention [48] and channel attention [48]. Specifically, as a typical design focusing on the feature channel dimension, channel attention [48] mainly conducts quantitative analysis on the importance of different feature channels, and generates attention weights in the channel dimension based on this analysis. It can not only enhance the feature response of key information channels but also suppress the interference of redundant channels, realizing accurate screening at the feature channel level. In contrast, spatial attention [48] focuses its optimization on the feature spatial dimension: by capturing the differences in semantic importance of different spatial positions in the feature map, it generates attention weights in the spatial dimension, thereby highlighting effective features in key spatial regions while weakening invalid information in background regions, and completing targeted enhancement at the feature spatial level. Although these two attention mechanisms focus on different dimensions of features respectively, both belong to the current mainstream efficient attention design patterns and have been proven to have practical value in feature optimization tasks. This ensures the rationality of conducting a fair comparison between them and (S)W-MTMSA.

**Tab. 7** Comparative experiment of spatial attention, channel attention and (S)W-MTMSA.

| Method | PLCC | SROCC | KROCC | Score |
|---|---|---|---|---|
| Spatial Attention [48] | 0.9480 | 0.9507 | 0.8515 | 2.7502 |
| Channel Attention [48] | 0.9472 | 0.9515 | 0.8556 | 2.7542 |
| **(S)W-MTMSA (Ours)** | **0.9507** | **0,9554** | **0.8629** | **2.7690** |

**Tab. 7** presents the comparison results between (S)W-MTMSA and the above two attention mechanisms, providing intuitive data support for the effectiveness of (S)W-MTMSA. From the experimental data in **Tab. 7**, it can be clearly observed that the proposed (S)W-MTMSA exhibits significant advantages in all key metrics for abdominal CT image quality assessment: whether it is PLCC (which reflects the linear correlation between predicted values and subjective scores), SROCC (which embodies the consistency of rank ordering), KROCC (which measures classification accuracy), or the sum of absolute values (Score) that integrates the evaluation capabilities of the three, the performance of (S)W-MTMSA in all metrics fully surpasses that of the other two comparative attention mechanisms. This fully proves the effectiveness and adaptability of (S)W-MTMSA in capturing the quality features of abdominal CT images, making it a more advantageous choice than the other two attention mechanisms in this specific task scenario.

#### *4.5.3. HRPS Loss*

To systematically verify the overall effectiveness of the proposed HRPS (High-Resolution Perceptual Similarity) loss function and the specific contributions of its internal core components to model performance, we designed a dedicated ablation experiment targeting the key components of this loss function. By removing different components one by one and comparing the changes in model performance, the functional value of each component is clarified. The experimental design and operation details are as follows: Experiment 1# (HRPS w/o Fine RPS): The fine RPS (Rank-Preserving Similarity) loss term in HRPS is removed independently, this component is mainly used to optimize the similarity matching of fine-scale features, and this experiment can verify the impact of fine feature constraints on model performance; Experiment 2# (HRPS w/o CrossEntropy): In the original structure of the HRPS loss function, only the cross-entropy loss term (which is responsible for optimizing the class probability distribution) is removed, while the remaining components are retained, and this experiment is intended to evaluate the necessity of this loss term in class discrimination tasks;

Experiment 3# (HRPS w/o Entropy Regularizer): The entropy regularization term is removed independently from the HRPS loss function, and this is to verify the contribution of this regularization component to the stability of model training; Experiment 4# (HRPS w/o Coarse RPS): Only the coarse RPS loss term in HRPS is removed—this component focuses on optimizing the rank consistency of global features, and through this experiment, we can analyze the role of coarse feature constraints in model training.

**Tab. 8** Ablation study of HRPS loss.

| Method | PLCC | SROCC | KROCC | Score |
|---|---|---|---|---|
| 1# HRPS w/o Fine RPS | 0.9484 | 0.9528 | 0.8552 | 2.7564 |
| 2# HRPS w/o CrossEntropy | 0.9482 | 0.9528 | 0.8555 | 2.7566 |
| 3# HRPS w/o Entropy Regularizer | 0.9481 | 0.9531 | 0.8567 | 2.7578 |
| 4# HRPS w/o Coarse RPS | 0.9489 | 0.9540 | 0.8583 | 2.7612 |
| **5# ClinReadNet (Ours)** | **0.9507** | **0.9554** | **0.8629** | **2.7690** |

From the ablation experiment results presented in **Tab. 8**, it can be clearly observed that each component of the HRPS loss function has a significant impact on model performance, and the complete HRPS loss function achieves the optimal performance, with an overall Score of 2.7690, and PLCC, SROCC, and KROCC of 0.9507, 0.9554, and 0.8629 respectively. This fully verifies the effectiveness of the overall design of the loss function. Specifically, the results of Experiment 1# (HRPS w/o Fine RPS) show that the model's Score decreases by 0.0126 to 2.7564; in Experiment 2# (HRPS w/o CrossEntropy), the model's Score decreases by 0.0124 compared to the complete HRPS, dropping to 2.7566; the results of Experiment 3# (HRPS w/o Entropy Regularizer) show that the model's Score decreases by 0.0112 to 2.7578; in Experiment 4# (HRPS w/o Coarse RPS), the model's Score drops to 2.7612, a decrease of 0.0078 compared to the complete function.

#### 4.5.4 Hyperparameters sensitivity analysis

In order to validate the stability and reliability of our proposed method, a sensitivity analysis was conducted on key hyperparameters.

We designed and carried out experiments to assess how variations in the initial values of adaptive weights for weighted residual connections within the SOQN module influence model performance—specifically by assigning distinct values to these initial weights.

As shown in **Figure 8**, the scores remain stable within a wide range of values for $\beta$ and $\gamma$. This phenomenon not only confirms that the method we proposed exhibits strong robustness against fluctuations in such hyperparameters, but also can be explained from the perspective of parameter characteristics: since the subsequent actual values of these hyperparameters are dynamically adjusted based on feedback results during the training process, the impact of their initial values on the final model performance is substantially reduced, and thus no obvious fluctuations in scores are caused.

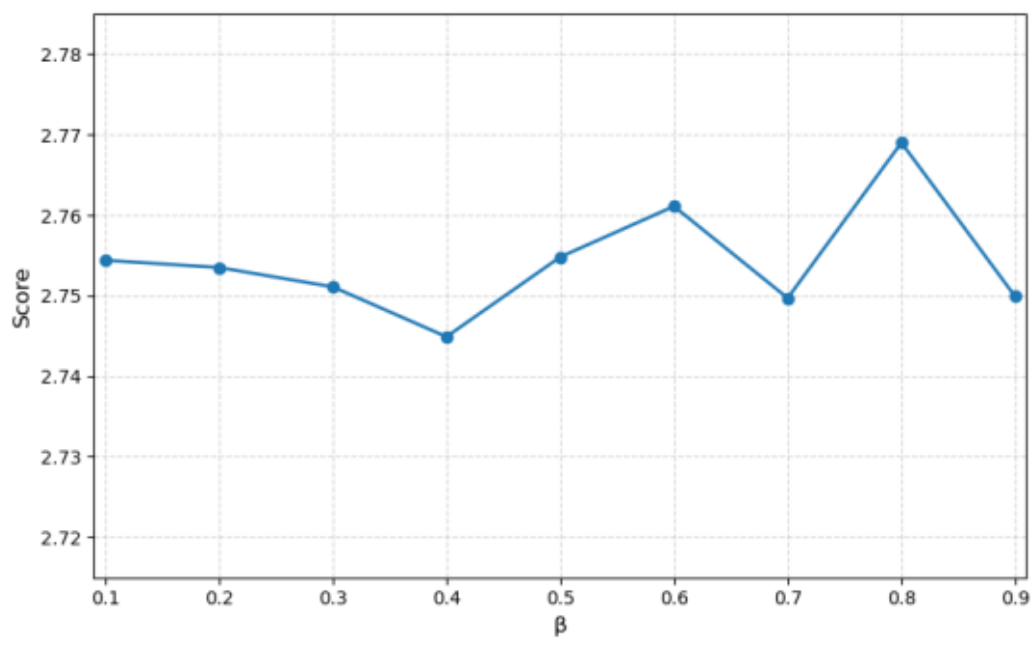

**Fig. 8** Visualization analysis of hyperparameters sensitivity for β and γ. β + γ =1.

To evaluate how changes in the weight of the loss term affect the performance of the model, we conducted experiments by taking different values of the weight of the loss term.

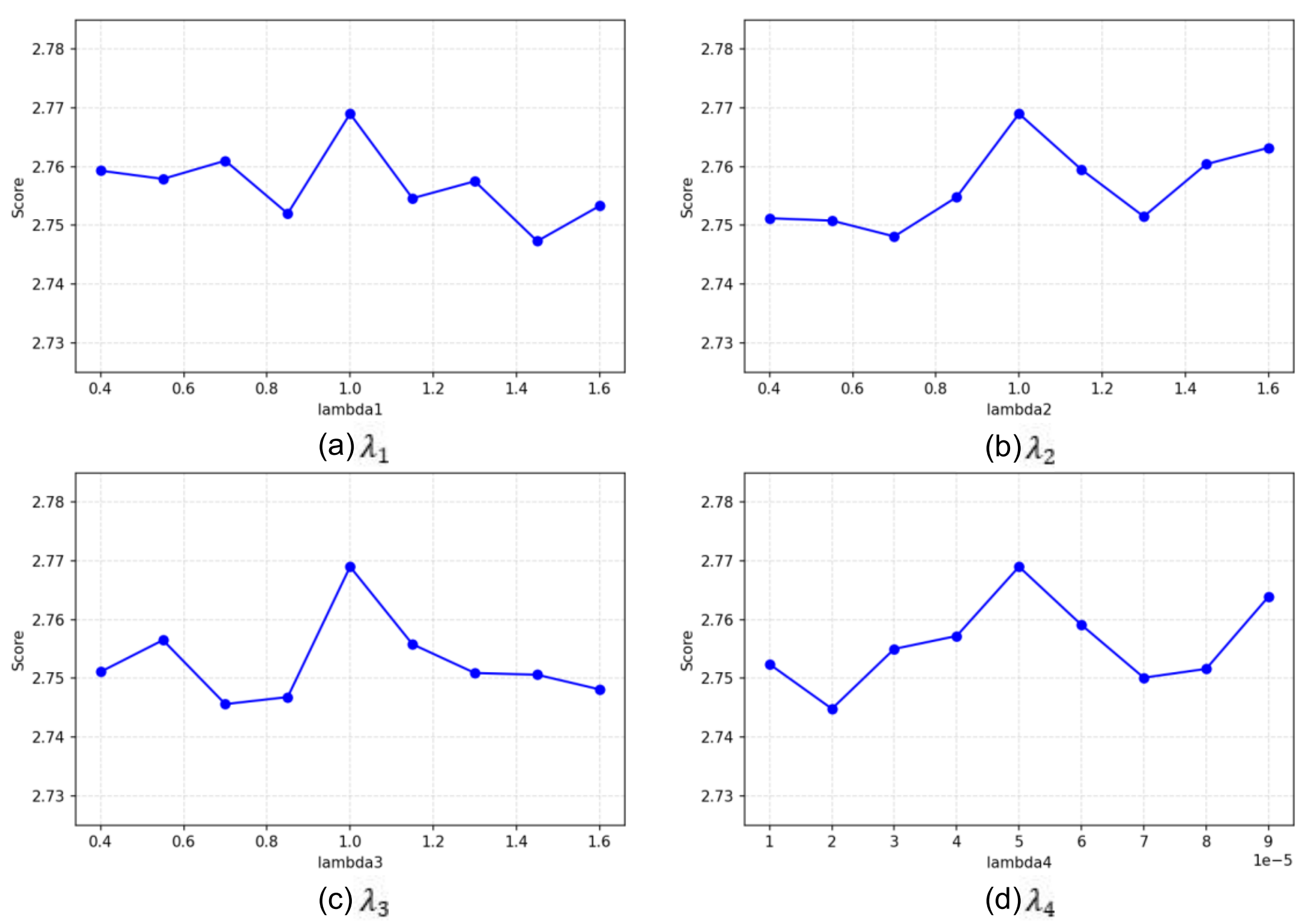

**Fig. 9** Visualization analysis of hyperparameters sensitivityfor $\lambda_1$, $\lambda_2$, $\lambda_3$, and $\lambda_4$.

As illustrated in **Fig. 9**, the Score remains stable across a broad range of values for $\lambda_1$, $\lambda_2$, $\lambda_3$, and $\lambda_4$—this observation validates the robustness of our method against fluctuations in these hyperparameters. It is noteworthy that the model achieves optimal performance when $\lambda_1$, $\lambda_2$, and $\lambda_3$ are all set to 1, which confirms the effectiveness of the MetaBalance gradient adjustment mechanism[39] we adopted for the task.

# 5. Discussion

## 5.1 The performance of ClinReadNet

This paper proposes an innovative framework that integrates the SOQN module, the (S)W-MTMSA module, and the HRPS loss.

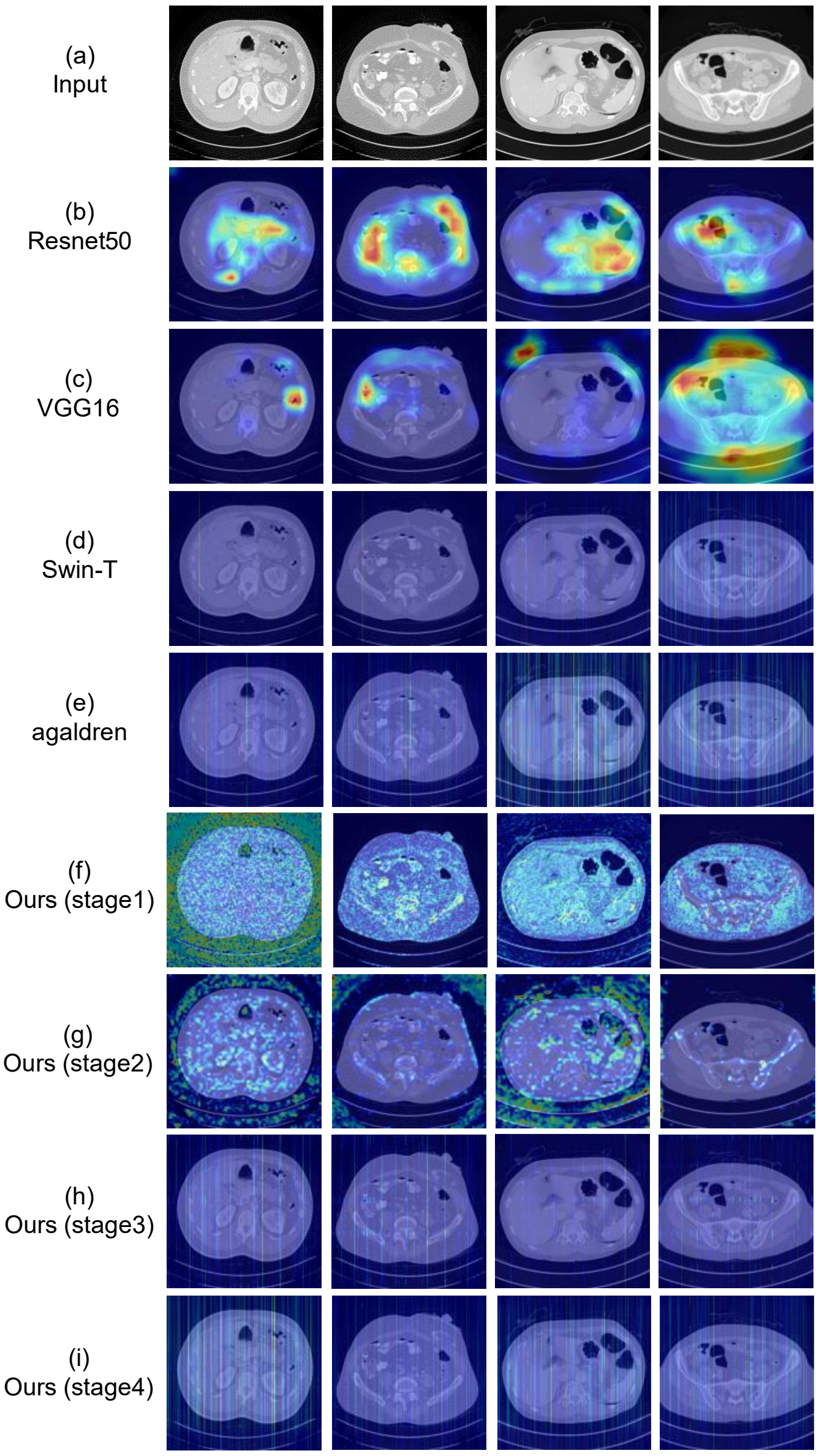

**Fig. 10** Heatmaps of CT images with different models.

As shown in **Tab. 2**, we conducted comparative experiments among general classification methods, the state-of-the-art (SOTA) methods from the challenge, and the method proposed in this paper. To further perform an intuitive and interpretable evaluation of the proposed model, we adopted the Grad-CAM method to generate heatmaps for the classic neural network-based model ResNet [44], the classic Transformer-based model Swin-T[26], the challenge's SOTA model agaldran [16], and our proposed model. These heatmaps can clearly reflect the regions that the models focus on during the prediction process.

As shown in **Fig. 10b** and **Fig. 10c**, ResNet50 [44] and VGG16 [46]—which are based on convolutional network architectures—concentrate their attention on the content information of organs, which does not fully align with the key focus of the task. In contrast, Swin-T[26] and agaldran[16] adopt a sparse sampling attention mode. They avoid interference from organ structures in the images through a stripe-like attention distribution and focus more on quality features related to human visual perception, such as the image noise level.

It can be observed from the heatmaps that the proposed ClinReadNet model integrates the advantages of the two aforementioned types of models. Our model uses Swin-T[26] as its basic backbone, and the performance of its heatmaps is shown in **Fig. 10f, 10g, 10h, 10i**: After the first two Swin blocks are processed by the SOQN module, they focus on capturing the edge features and global quality features of the image, enhancing the attention to the structural information that is core to the task; the third Swin Block retains the sparsely sampled feature maps, maintaining the transformer architecture's ability to perceive unstructured quality factors; the fourth Swin block generates sparsely sampled feature maps through a multi-temperature mechanism, further optimizing the selective focus on quality-related cues. Finally, the feature maps output from top to bottom by the four Swin Blocks are concatenated and fused. This not only absorbs the advantage of convolutional networks in accurately focusing on image content and guides it to the edge feature level that is more suitable for the quality assessment task but also retains the Transformer architecture's ability to sensitively capture unstructured quality factors through sparse sampling. As a result, a single model can meet the dual requirements of structural details and global performance in CT image quality assessment, leading to a significant improvement in the discrimination accuracy and adaptability in complex quality scenarios.

## 5.2 Ablation Study

In Section 4.5, we have verified the effectiveness of each proposed method through ablation experiments. To further deepen the understanding of the roles of the three key components—the SOQN, (S)W-MTMSA, and HRPS loss function—this section employs visualization techniques to conduct a more intuitive analysis of their functions.

### *5.2.1. The function of* **SOQN**

With the synergy of its dual branches, the SOQN module not only accurately captures microstructures and enhances sensitivity to local defects via the Sobel Branch but also models overall quality through the Ordinal Quality Branch, thereby achieving comprehensive and precise assessment of LDCT image quality. As shown in **Tab. 4**, we verified the effectiveness of the module and the advantage of the two branches working in synergy through ablation experiments. To explain the function of SOQN more intuitively, we conducted visualization analysis on the feature maps before and after processing by the network. **Fig. 11** presents the feature maps before and after processing by SOQN1 and SOQN2, respectively. The purpose of this visualization analysis is twofold: first, to clearly demonstrate the specific process by which SOQN processes and enhances image features; second, to empirically verify the negative impact on model performance when only one branch of SOQN is used independently.

**Fig. 11a** shows the feature maps before and after processing by SOQN1 and SOQN2. In the feature maps before SOQN module processing, the content they focus on is relatively scattered and sparse. In contrast, in the feature maps processed by these two SOQN modules, edge information and detailed information are effectively enhanced, while attention to the background—relevant to hierarchical quality—is increased. This feature performance is more conducive to the quality assessment of CT images.

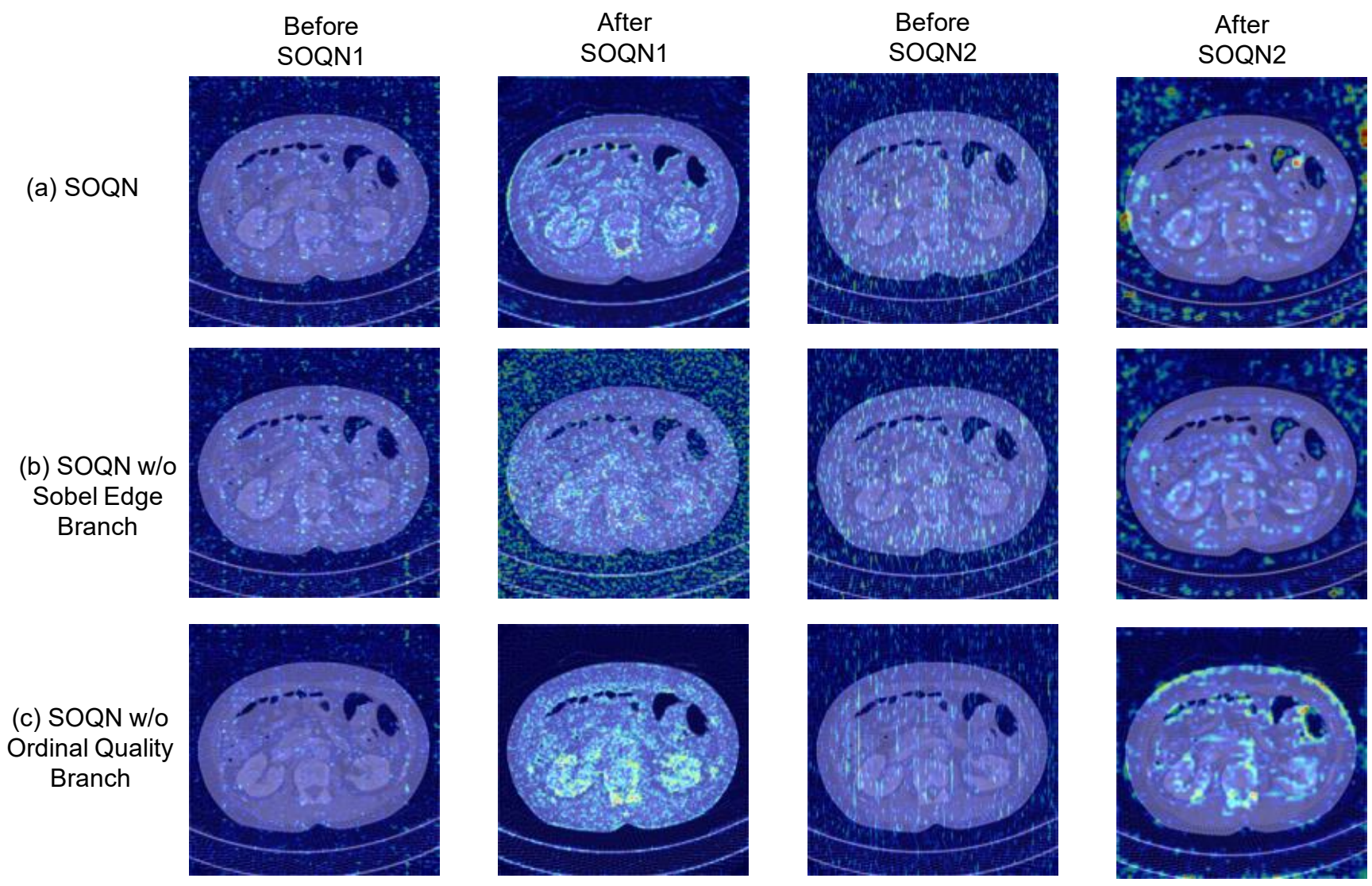

**Fig. 11** Visualization of feature maps before and after the SOQN1(SOQN2).

On the other hand, **Fig. 11b** and **Fig. 11c** present the results when only the Ordinal Quality Branch or the Sobel Branch is used independently. Compared with the output of the complete SOQN module in **Fig. 11a**, each of these two modules has its own focus. As shown in **Fig. 11b**, the Ordinal Quality Branch pays more attention to image features related to quality—not only including internal organ features but also background features. As shown in **Fig. 11c**, the Sobel Branch focuses more on edge features closely related to image quality, including organ edges and internal organ edges. Only the SOQN module that fuses the two branches can achieve better focus on image quality: it can not only capture edge features directly related to image quality but also attend to other quality-related features such as background features and internal organ features.

### *5.2.2. The function of (S)W-MTMSA*

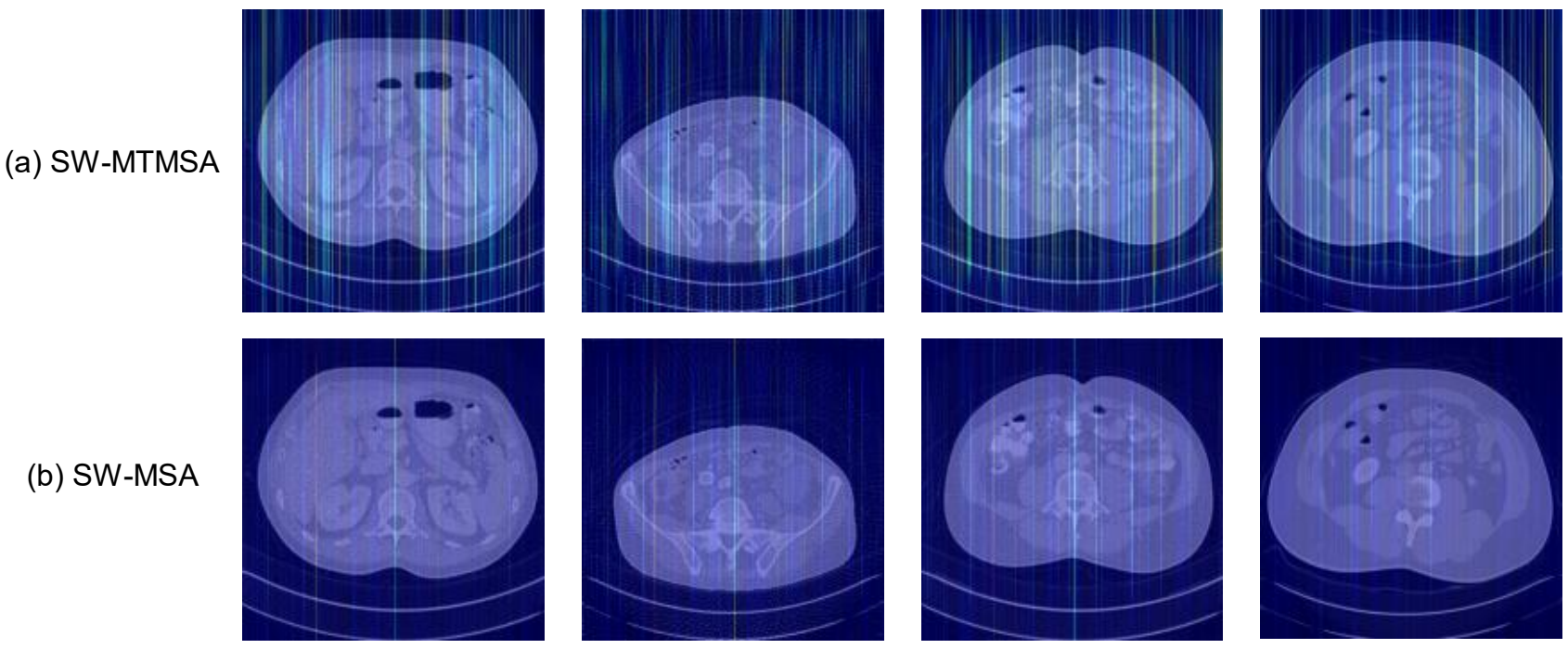

**Fig. 12** Heatmaps of CT images with (S)W-MTMSA and (S)W-MSA.

The (S)W-MTMSA module captures multi-granularity quality features through multi-temperature attention and adaptively adjusts weights via a quality-aware modulator. This not only aligns with the clinical observation logic of radiologists but also covers evaluation requirements in all dimensions and accurately adapts to the unique quality issues of individual images. As shown in **Tab. 6**, we verified the effectiveness of the (S)W-MTMSA module through ablation experiments. To more intuitively demonstrate the effect of the (S)W-MTMSA module, we compared it with the original attention mechanism (SW-MSA) in Swin-T and presented the comparison in the form of heatmaps for visualization.

As shown in **Fig. 12**, the sparse attention distribution of the original SW-MSA attention mechanism in Swin-T is relatively uniform, leading to insufficient differentiation of key factors affecting image quality. In contrast, the sparse attention of the (S)W-MTMSA module exhibits more obvious differences in regional focus. With a stripe-like attention pattern, it can avoid interference from organ structures while focusing more accurately on the core factors that affect image quality. The distinction in the depth of attention to different regions is more significant, thereby enhancing the pertinence of abdominal CT image quality assessment.

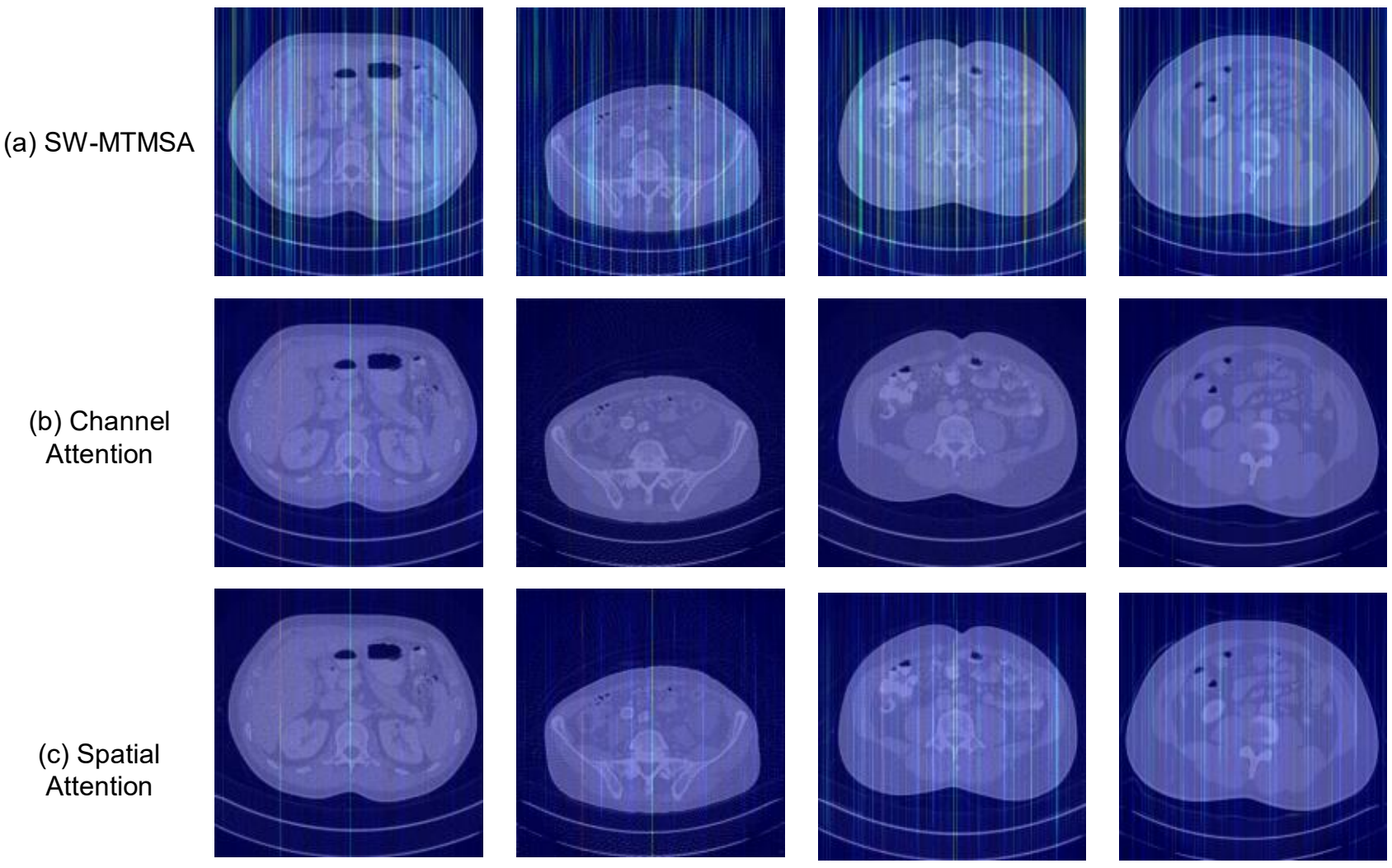

**Fig. 13** Visualization Analysis of (S)W-MTMSA, channel attention and spatial attention.

As shown in **Tab. 7**, we have verified that in the specific task of this study, the performance of (S)W-MTMSA is superior to that of channel attention [48] and spatial attention [48]. To more intuitively present the differences between them, we used heatmaps to visualize the attention maps generated by the two attention mechanisms in the model.

As demonstrated in **Fig. 13b**, the attention distribution of channel attention is relatively uniform, with insufficiently significant differences in attention to different regions, making it difficult to accurately focus on key quality features. As shown in **Fig. 13c**, although spatial attention focuses more on the foreground regions of the image compared to channel attention and shows a certain degree of distinction in the depth of attention to different regions, its coverage range is relatively limited. In contrast, (S)W-MTMSA—as shown in **Fig. 13a**—adopts a multi-region, differentiated sparse stripe attention pattern. It can not only cover quality-related features in multiple locations of the image (such as organ edges and potential artifacts) simultaneously but also accurately distinguish the attention intensity of different regions. While avoiding excessive interference from organ structures, it also takes into account the perception of local details and global trends, which is more in line with the clinical observation logic of radiologists and demonstrates obvious advantages in abdominal CT image quality assessment.

#### *5.2.3. The function of HRPS Loss*

The HRPS loss function adapts to clinical scenarios through a two-layer ranked probability score, models the ordered relationship of quality grades to align with doctors' judgments, balances loss scales via gradient adjustment mechanism [39], and mitigates overfitting with entropy regularization [40]—comprehensively optimizing model training. The synergy of these components enables the framework to achieve more accurate LDCT image quality evaluation that better meets clinical needs.

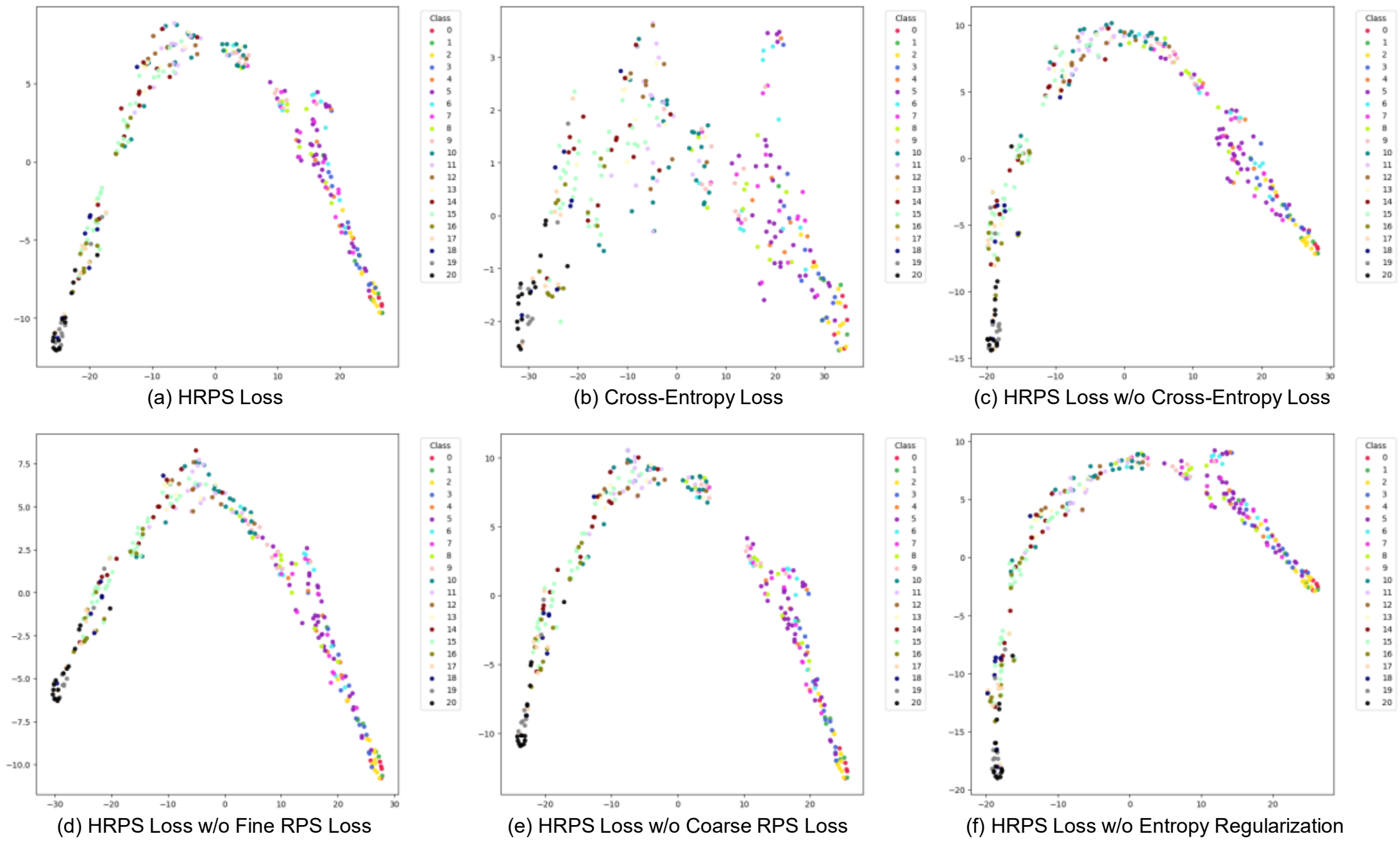

**Fig. 14** Visualization analysis of the components of HRPS loss.

As shown in **Tab. 8**, the ablation experiment results confirm that the HRPS loss function exhibits excellent performance in this task, and each of its components is irreplaceable. To further explore the internal mechanism behind its performance improvement, we extracted the 2D features from the last hidden layer of the test set and presented the feature distribution in the form of scatter plots, thereby intuitively analyzing the impact of different components on image quality assessment.

The role of each loss component can be intuitively determined from the scatter distribution in the figures: In **Fig. 14b** (single cross-entropy loss), the points are scattered and disorganized, indicating that the model has weak ability to distinguish images with ordered quality grades. In contrast, the scatter plots of other combinations show that the points exhibit a regular continuous trend and points of the same category are more clustered rather than scattered; however, compared with the complete HRPS loss function, each of these combinations has its own shortcomings. For example, in **Fig. 14c** (HRPS w/o cross-entropy loss), the points lack aggregation, which proves that the cross-entropy loss term is crucial for the basic category distinction of quality images. In **Fig. 14d** (HRPS w/o fine RPS loss), although the distribution shows a certain pattern, the discrimination of fine quality grades is insufficient—highlighting the importance of fine RPS loss for the refined evaluation of quality scores. In **Fig. 14e** (HRPS w/o coarse RPS loss), the model's ability to distinguish overall quality trends decreases, demonstrating that coarse RPS loss is indispensable for constructing the macro ordered relationship of quality grades. In **Fig. 14f** (HRPS w/o entropy regularization term), the discrimination of scatter points decreases, underscoring the role of the entropy regularization term in ensuring discrimination and mitigating overfitting.

In contrast, **Fig. 14a** (complete HRPS loss) shows that the points exhibit a highly aggregated and ordered distribution pattern. Thanks to the synergy of all components, the model not only possesses the ability for basic category distinction but also achieves refined evaluation of quality scores, constructs the macro ordered relationship of quality grades, and avoids overfitting. When distinguishing images of different qualities, it aligns with the logical thinking of doctors' subjective judgments, can accurately and stably cover the full-dimensional quality assessment needs, and demonstrates comprehensive advantages that models with single components or missing components do not have.

### 5.3 Limitations and Future work

Despite its promising performance, the proposed ClinReadNet has certain limitations that outline the boundaries of its current contributions and point to future research directions.

1. The current model operates on 2D slices. Although this is a common practice, it does not fully leverage the 3D contextual information inherent in volumetric CT data. A promising future direction is to extend the clinical-reading-inspired paradigm to 3D, developing mechanisms that can simulate a radiologist's scrolling through adjacent slices to assess quality comprehensively.

2. The interpretability of the model, while improved through clinical inspiration, could be further quantitatively evaluated. For instance, conducting eye-tracking studies to compare the attention maps of ClinReadNet with the gaze patterns of radiologists could provide more rigorous, human-centric validation of the model's decision-making process.

## 6. Conclusion

This study proposes the ClinReadNet model inspired by clinical image reading and leveraging the ordered nature of task labels. The model designs the SOQN module, which adopts a dual-branch structure to focus on edge features highly relevant to image quality, exploit the ordered nature of labels, and fully integrate local and global information to enhance model performance. The (S)W-MTMSA module is devised, drawing inspiration from clinical image reading to mimic the process of doctors shifting from focusing on the overall picture to details. It fuses feature maps with different sharpness levels through adaptive weights, improving the model's quality detection capability. Meanwhile, an innovative HRPS loss function is proposed, which makes full use of the ordered nature of the task through the RPS loss, which tailored for ordered problems, for both coarse and fine classification. It also aligns with clinical image reading, making prediction results closer to the ground truth. Ablation experiments verify the effectiveness of each component. Our model achieves state-of-the-art (SOTA) performance, with a Score of 2.7690 on the LDCTIQAG2023 dataset, and PLCC, SROCC, and KROCC reaching 0.9507, 0.9554, and 0.8629, respectively. We believe that ClinReadNet can provide valuable support for doctors in their abdominal CT image quality assessment work.

**Declaration of competing interest**

The authors declare that there are no conflicts of interest regarding the publication of this paper.

**CRediT authorship contribution statement**

**Xianye Xiao**: Conceptualization; Data curation; Formal analysis; Investigation; Methodology; Software; Validation; Visualization; Writing-original draft. **Yulong Zou**: Software; Investigation; Validation; Visualization; Writing-review & editing. **Yujie Luo**: Investigation; Validation; Writing-review & editing. **Taihui Yu**: Investigation; Validation; Writing-review & editing. **Cun-Jing**

**Zheng**: Investigation; Validation; Writing-review & editing. **Yuan-ming Geng**: Investigation; Validation; Writing-review & editing. **Shuihua Wang**: Validation, Writing-review & editing. **Yudong Zhang**: Validation, Writing-review & editing. **Jin Hong**: Conceptualization; Data curation; Investigation; Methodology; Resources; Funding acquisition; Project administration; Supervision; Writing-review & editing, Writing-original draft.

**Acknowledgements**

This work was supported in part by the National Natural Science Foundation of China (62466033), and in part by the Jiangxi Provincial Natural Science Foundation (20242BAB20070).

## References

1. Sadia, R.T., J. Chen, and J. Zhang, *CT image denoising methods for image quality improvement and radiation dose reduction.* Journal of applied clinical medical physics, 2024. **25**(2): p. e14270.
2. Sodickson, A., et al., *Recurrent CT, cumulative radiation exposure, and associated radiation-induced cancer risks from CT of adults.* Radiology, 2009. **251**(1): p. 175-184.
3. Slaney, M. and A. Kak, *Principles of computerized tomographic imaging*. 1988: IEEE press.
4. Chow, L.S., et al., *Correlation between subjective and objective assessment of magnetic resonance (MR) images.* Magnetic resonance imaging, 2016. **34**(6): p. 820-831.
5. Chow, L.S. and R. Paramesran, *Review of medical image quality assessment.* Biomedical signal processing and control, 2016. **27**: p. 145-154.
6. Wang, Z. and A.C. Bovik, *Modern image quality assessment.* 2006.
7. Wang, Z., et al., *Image quality assessment: from error visibility to structural similarity.* IEEE transactions on image processing, 2004. **13**(4): p. 600-612.
8. Mittal, A., R. Soundararajan, and A.C. Bovik, *Making a "completely blind" image quality analyzer.* IEEE Signal processing letters, 2012. **20**(3): p. 209-212.
9. Mittal, A., A.K. Moorthy, and A.C. Bovik, *No-reference image quality assessment in the spatial domain.* IEEE Transactions on image processing, 2012. **21**(12): p. 4695-4708.
10. Lee, W., et al., *No-reference perceptual CT image quality assessment based on a self-supervised learning framework.* Machine Learning: Science and Technology, 2022. **3**(4): p. 045033.
11. Mason, A., et al., *Comparison of objective image quality metrics to expert radiologists' scoring of diagnostic quality of MR images.* IEEE transactions on medical imaging, 2019. **39**(4): p. 1064-1072.
12. Liu, B., et al., *DGSSA: Domain generalization with structural and stylistic augmentation for retinal vessel segmentation.* Neural Networks, 2026. **194**: p. 108118.
13. Cheng, J., et al., *WaveNet-SF: A hybrid network for retinal disease detection based on wavelet transform in spatial-frequency domain.* Neural Networks, 2026. **194**: p. 108189.
14. Wan, Z., et al., *Data generation for enhancing EEG-based emotion recognition: Extracting time-invariant and subject-invariant components with contrastive learning.* IEEE Transactions on Consumer Electronics, 2024.
15. Gao, L., et al., *Autism spectrum disorders detection based on multi-task transformer neural network.* BMC neuroscience, 2024. **25**(1): p. 27.

16. Lee, W., et al., *Low-dose computed tomography perceptual image quality assessment.* Medical Image Analysis, 2025. **99**: p. 103343.
17. Huang, Y., et al. *Lesion-based contrastive learning for diabetic retinopathy grading from fundus images*. in *International Conference on Medical Image Computing and Computer-Assisted Intervention*. 2021. Springer.
18. Cheng, J., *Sparse range-constrained learning and its application for medical image grading.* IEEE transactions on medical imaging, 2018. **37**(12): p. 2729-2738.
19. Khawaldeh, S., et al., *Noninvasive grading of glioma tumor using magnetic resonance imaging with convolutional neural networks.* Applied Sciences, 2017. **8**(1): p. 27.
20. Shazuli, S.I.S.M. and A. Saravanan, *Improved whale optimization algorithm with deep learning-driven retinal fundus image grading and retrieval.* Engineering, Technology & Applied Science Research, 2023. **13**(5): p. 11555-11560.
21. Han, M. and J. Baek, *A convolutional neural network-based anthropomorphic model observer for signal-known-statistically and background-known-statistically detection tasks.* Physics in Medicine & Biology, 2020. **65**(22): p. 225025.
22. Lauermann, J., et al., *Automated OCT angiography image quality assessment using a deep learning algorithm.* Graefe's Archive for Clinical and Experimental Ophthalmology, 2019. **257**(8): p. 1641-1648.
23. Gao, Q., et al. *Blind CT image quality assessment via deep learning framework*. in *2019 IEEE Nuclear Science Symposium and Medical Imaging Conference (NSS/MIC)*. 2019. IEEE.
24. Qi, C., et al., *An artificial intelligence-driven image quality assessment system for whole-body [18F] FDG PET/CT.* European Journal of Nuclear Medicine and Molecular Imaging, 2023. **50**(5): p. 1318-1328.
25. Tan, M. and Q. Le. *Efficientnetv2: Smaller models and faster training*. in *International conference on machine learning*. 2021. PMLR.
26. Liu, Z., et al. *Swin transformer: Hierarchical vision transformer using shifted windows*. in *Proceedings of the IEEE/CVF international conference on computer vision*. 2021.
27. Kolesnikov, A., et al. *Big transfer (bit): General visual representation learning*. in *European conference on computer vision*. 2020. Springer.
28. Galdran, A., et al. *Multi-head multi-loss model calibration*. in *International conference on medical image computing and computer-assisted intervention*. 2023. Springer.
29. Epstein, E.S., *A scoring system for probability forecasts of ranked categories.* Journal of Applied Meteorology (1962-1982), 1969. **8**(6): p. 985-987.
30. Galdran, A. *Performance metrics for probabilistic ordinal classifiers*. in *International Conference on Medical Image Computing and Computer-Assisted Intervention*. 2023. Springer.
31. Yang, S., et al. *Maniqa: Multi-dimension attention network for no-reference image quality assessment*. in *Proceedings of the IEEE/CVF conference on computer vision and pattern recognition*. 2022.
32. Dosovitskiy, A., et al., *An image is worth 16x16 words: Transformers for image recognition at scale.* arXiv preprint arXiv:2010.11929, 2020.
33. Löfstedt, T., et al., *Gray-level invariant Haralick texture features.* PloS one, 2019. **14**(2): p. e0212110.

34. Cohen, J., *Weighted kappa: Nominal scale agreement provision for scaled disagreement or partial credit.* Psychological bulletin, 1968. **70**(4): p. 213.
35. Ferrer, L., *Analysis and comparison of classification metrics.* arXiv preprint arXiv:2209.05355, 2022.
36. Gneiting, T. and A.E. Raftery, *Strictly proper scoring rules, prediction, and estimation.* Journal of the American statistical Association, 2007. **102**(477): p. 359-378.
37. Maier-Hein, L. and B. Menze, *Metrics reloaded: Pitfalls and recommendations for image analysis validation.* arXiv. org, 2022(2206.01653).
38. Jang, E., S. Gu, and B. Poole, *Categorical reparameterization with gumbel-softmax.* arXiv preprint arXiv:1611.01144, 2016.
39. He, Y., et al. *Metabalance: improving multi-task recommendations via adapting gradient magnitudes of auxiliary tasks*. in *Proceedings of the ACM Web Conference 2022*. 2022.
40. Ziebart, B.D., et al. *Maximum entropy inverse reinforcement learning*. in *Aaai*. 2008. Chicago, IL, USA.
41. Xun, S., et al., *Charting the path forward: CT image quality assessment-an in-depth review.* Journal of King Saud University Computer and Information Sciences, 2025. **37**(5): p. 1-24.
42. Tan, M. and Q. Le, *Efficientnetv2: Smaller models and faster training. arXiv 2021.* arXiv preprint arXiv:2104.00298, 2021. **5**.
43. Sandler, M., et al. *Mobilenetv2: Inverted residuals and linear bottlenecks*. in *Proceedings of the IEEE conference on computer vision and pattern recognition*. 2018.
44. He, K., et al. *Deep residual learning for image recognition*. in *Proceedings of the IEEE conference on computer vision and pattern recognition*. 2016.
45. Liu, Z., et al. *A convnet for the 2020s*. in *Proceedings of the IEEE/CVF conference on computer vision and pattern recognition*. 2022.
46. Simonyan, K. and A. Zisserman, *Very deep convolutional networks for large-scale image recognition.* arXiv preprint arXiv:1409.1556, 2014.
47. Bao, H., et al., *Beit: Bert pre-training of image transformers.* arXiv preprint arXiv:2106.08254, 2021.
48. Park, J., et al., *Bam: Bottleneck attention module.* arXiv preprint arXiv:1807.06514, 2018.